\documentclass[10pt,twocolumn,letterpaper]{article}

\usepackage[pagenumbers]{cvpr} 

\usepackage{graphicx}
\usepackage{amsmath}
\usepackage{amssymb}
\usepackage{booktabs}

\usepackage{epsfig}
\usepackage{graphicx}
\usepackage{enumitem}
\usepackage{bbding} 

\usepackage{times}
\usepackage{caption}
\usepackage{subcaption}
\usepackage{amsmath}
\usepackage{array}
\usepackage{colortbl}
\usepackage{bbm}
\usepackage{multirow}
\usepackage{multicol}
\usepackage{makecell}
\usepackage{xcolor}
\usepackage{xspace}
\usepackage{float}
\usepackage{color}

\usepackage{pifont}

\usepackage{booktabs}
\usepackage{tabulary}
\usepackage{dblfloatfix}
\usepackage{transparent}

\usepackage{algorithm}
\usepackage{listings}
\usepackage{algorithmic}
\usepackage{balance}
\usepackage{amssymb}

\usepackage{color, colortbl}
\definecolor{iGray}{gray}{0.9}
\definecolor{beaublue}{rgb}{0.74, 0.83, 0.9}
\definecolor{Royal_Blue}{rgb}{0.0, 0.1, 0.66}
\usepackage[pagebackref=true,breaklinks=true,letterpaper=true,colorlinks,bookmarks=false, citecolor = Royal_Blue]{hyperref}

\usepackage{pifont}

\usepackage{bbding}
\usepackage{threeparttable}
\usepackage{multirow}
\usepackage{longtable}
\usepackage{rotating}
\usepackage{tabularx}
\usepackage{graphicx}
\usepackage{mwe}
\usepackage{amsmath, bm}
\usepackage{float}
\usepackage{caption}
\usepackage{pifont}
\usepackage{bbding}

\usepackage{graphicx}
\usepackage{mwe}
\usepackage{amsmath, bm}
\usepackage{float}
\usepackage{caption}

\usepackage{graphicx}
\usepackage{amsfonts}
\usepackage{color}
\usepackage{bm}
\usepackage{amsmath}
\usepackage{amssymb}
\usepackage{multirow}
\usepackage{makecell}
\usepackage{enumitem}
\usepackage{tabularx}
\usepackage{fixltx2e} 
\usepackage{booktabs} 
\usepackage{xcolor}
\usepackage{url}
\usepackage{graphicx}
\usepackage{amsfonts}
\usepackage{color}
\usepackage{bm}

\usepackage[capitalize]{cleveref}
\crefname{section}{Sec.}{Secs.}
\Crefname{section}{Section}{Sections}
\Crefname{table}{Table}{Tables}
\crefname{table}{Tab.}{Tabs.}

\def\cvprPaperID{3860} 
\def\confName{CVPR}
\def\confYear{2022}

\begin{document}

	\title{Correlation-Aware Deep Tracking}
	
	\author{Fei Xie$^{~\dagger}$\textsuperscript{\thanks {Work performed when Fei Xie was an intern of MSRA} }, Chunyu Wang$^{~\ddagger}$, Guangting Wang$^{~\ddagger}$, Yue Cao$^{~\ddagger}$, Wankou Yang$^{~\dagger}$,Wenjun Zeng$^{~\ddagger}$\\
		$^{\dagger}$~Southeast University, China\\$^{\ddagger}$~Microsoft Research Asia\\
		{\tt\small jaffe0319@gmail.com, chnuwa@microsoft.com, flylight@mail.ustc.edu.cn}\\
		{\tt\small   yuecao@microsoft.com, wkyang@seu.edu.cn, wezeng@microsoft.com }\\
	}

	\maketitle
	
	\begin{abstract}
		
		Robustness and discrimination power are two fundamental requirements in visual object tracking. 
		In most tracking paradigms, we find that the features extracted by the popular Siamese-like networks cannot fully discriminatively model the tracked targets and distractor objects, hindering them from simultaneously meeting these two requirements.
		While most methods focus on designing robust correlation operations, we propose a novel target-dependent feature network inspired by the self-/cross-attention scheme. 
		In contrast to the Siamese-like feature extraction,
		our network deeply embeds cross-image feature correlation in multiple layers of the feature network. 
		By extensively matching the features of the two images through multiple layers, it is able to suppress non-target features, resulting in instance-varying feature extraction.
		The output features of the search image can be directly used for predicting target locations without extra correlation step.
		Moreover, our model can be flexibly pre-trained on abundant unpaired images, leading to notably faster convergence than the existing methods.
		Extensive experiments show our method achieves the state-of-the-art results while running at real-time. Our feature networks also can be applied to existing tracking pipelines seamlessly to raise the tracking performance. %
		Code will be available.
		
	\end{abstract}
	
	\section{Introduction}
	
	\begin{figure}[t]
		\centering{\includegraphics[scale = 0.35]{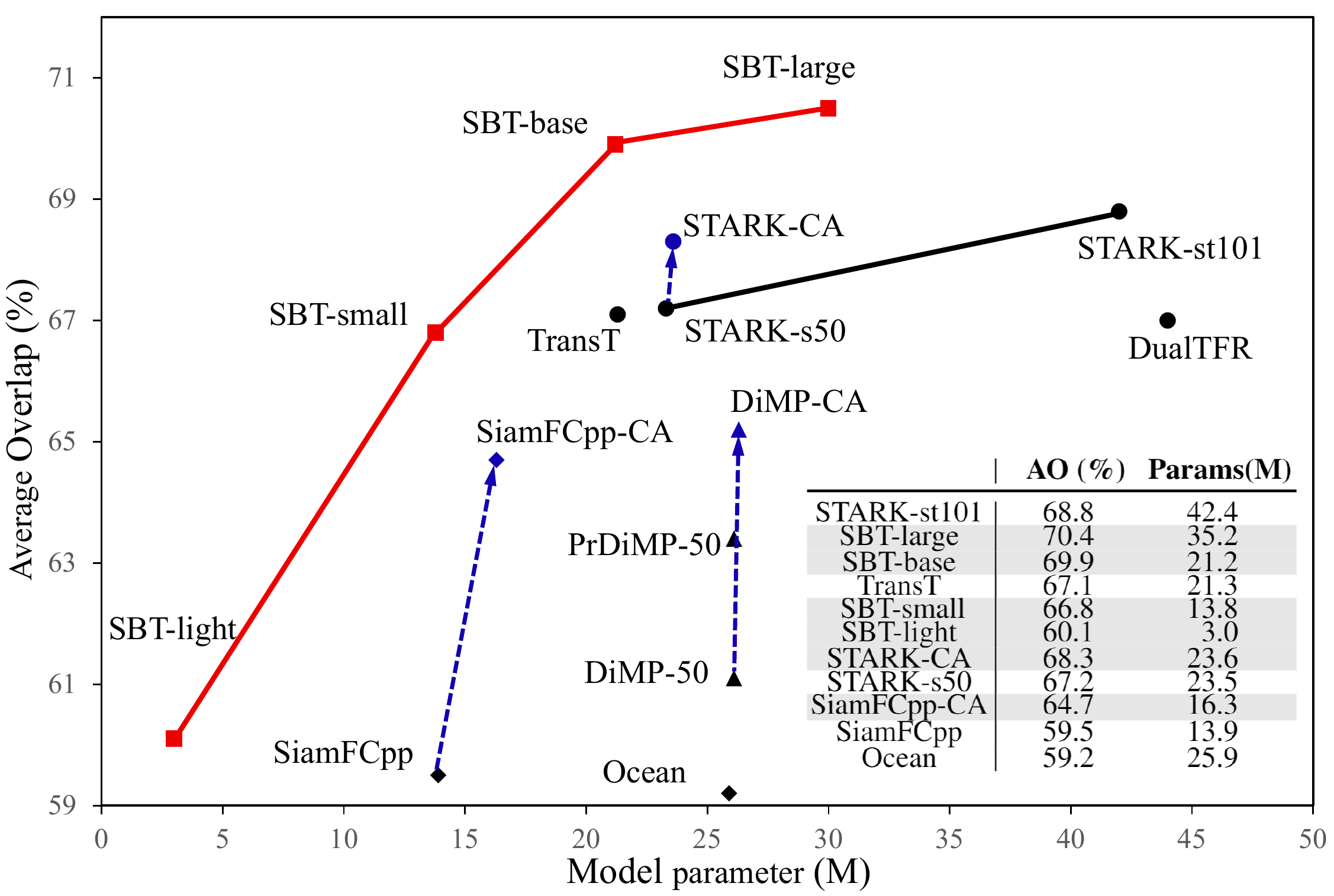}}
		\vspace{-1.0em}
		\caption{Comparison with the state-of-the-arts on  GOT-10k~\cite{GOT10K}. We visualize the AO performance with respect to the model size. All reported trackers follow the official GOT-10k test protocol. Our SBT tracker achieves superior results while multiple trackers (with suffix ``CA'') can benefit from our correlation-aware features.} 
		\vspace{-1.8em}
		\label{fig:bab}
	\end{figure}

	Visual object tracking (VOT) is a long-standing topic in computer vision. 
	There are two fundamental yet competing goals in VOT: on one hand, it needs to recognize the target undergoing large appearance variations; on the other hand, it needs to filter out the distractors in the background which may be very similar to the target.

	\begin{figure*}
		\begin{minipage}[t]{0.51\linewidth}
			\centering
			\includegraphics[scale = 0.42]{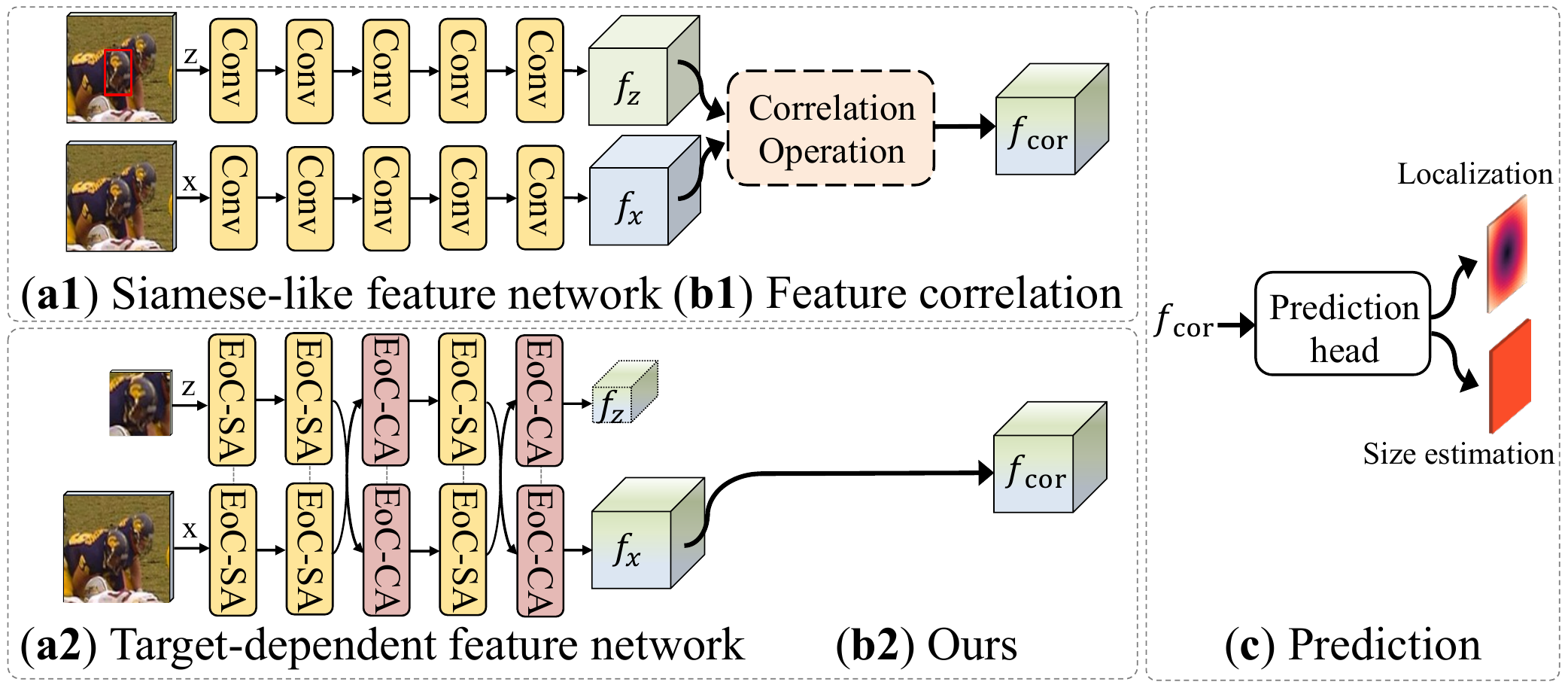}
		\end{minipage}%
		\begin{minipage}[t]{0.49\linewidth}
			\centering
			\includegraphics[scale = 0.40]{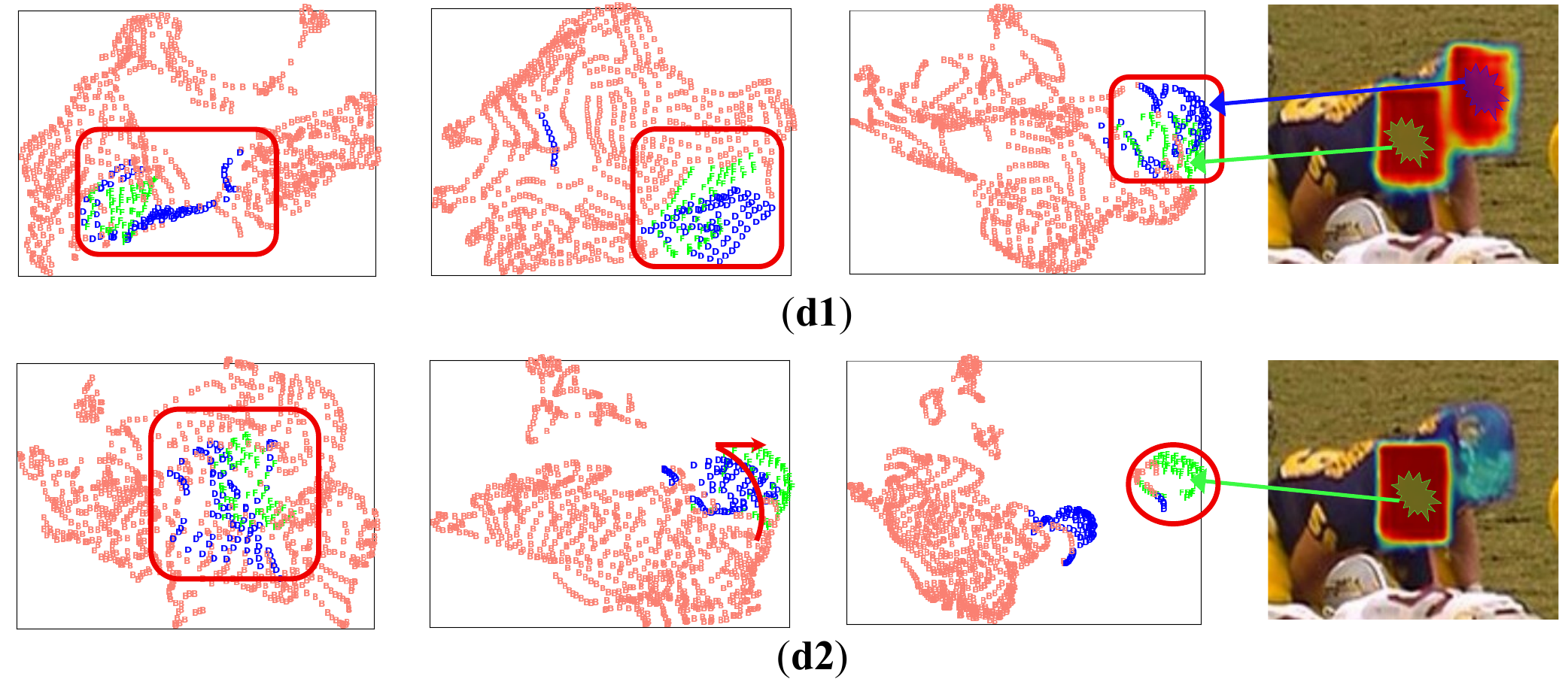}
		\end{minipage}
		
		\vspace{-0.8em}
		\caption{
			\textbf{(a1)} standard Siamese-like feature extraction;
			\textbf{(a2)} our target-dependent feature extraction;
			\textbf{(b1)} correlation step, such as Siamese cropping correlation~\cite{ siamrpn},  DCF~\cite{atom} and Transformer-based correlation~\cite{transt} ;
			\textbf{(b2)} our pipeline removes separated correlation step;
			\textbf{(c)} prediction stage;
			\textbf{(d1)}/\textbf{(d2)} are the TSNE~\cite{tsne} visualizations of search features in \textbf{(a1)}/\textbf{(a2)} when feature networks go deeper. 
		}
		\label{fig:CAcompare}
		\vspace{-1.5em}
	\end{figure*}

	Most appearance-based approaches address this challenge in two perspectives: the first is to learn a more expressive feature embedding space by Siamese-like extraction network~\cite{siamdw, siamrpn++}; the second is to develop a more robust correlation operation, such as Siamese cropping~\cite{siamrpn, siamdw}, online filter learning~\cite{KCF, DiMP} and Transformer-based fusion~\cite{transt, stark}.  
	Since the modern backbones~\cite{googlenet, ResNet} become the mainstream choice in deep era, most trackers devote to the correlation operation, hoping to discriminate targets from distractors given their features.
	Despite their great success, few of these tracking paradigms notice that the two competing goals may put the feature network into a target-distractor dilemma, bringing much difficulties to the correlation step. 
	The underlying reasons are three folds: 
	$1)$ The Siamese encoding process is unaware of the template and search images, which weakens the instance-level discrimination of learned embeddings.
	$2)$ There is no explicit modelling for the backbone to learn the decision boundary that separates the two competing goals, leading to a sub-optimal embedding space.
	$3)$ Each training video only annotates one single object while arbitrary objects including distractors can be tracked during inference. This gap is further enlarged by $2)$.
	Our key insight is that feature extraction should have dynamic instance-varying behaviors to generate “appropriate" embeddings for VOT to ease the dilemma.
	In more details, it needs to generate \emph{coherent} features for the same object in all frames of a video in spite of the variations;
	on the other hand, it needs to generate \emph{contrasting} features for the target and distractors with similar appearance.

	To this end, we present a novel dynamic feature network on top of the attention scheme~\cite{vaswani2017attention}. 
	As shown in Fig.\ref{fig:CAcompare} (a2), our Single Branch Transformer (SBT) network allows the features of the two images to deeply interact with each other at the stage of feature extraction.
	Intuitively,  the cross-attention weights gradually filter out target-irrelevant features layer by layer while the self-attention weights enrich the feature representations for better matching.
	Thus, the feature extraction process is target-dependent and asymmetrical for image pair, allowing the network to achieve a win-win scenario: it differentiates the target from similar distractors while preserving the coherent characteristics among dissimilar targets. The effectiveness of features from SBT is validated in Fig. \ref{fig:CAcompare} (d2). 
	The features belonging to the target (green) become more and more separated from the background (pink) and distractors (blue) while the search features from Siamese extraction are totally target-unaware.

\begin{figure*}[t]
		\centering{\includegraphics[scale = 0.5]{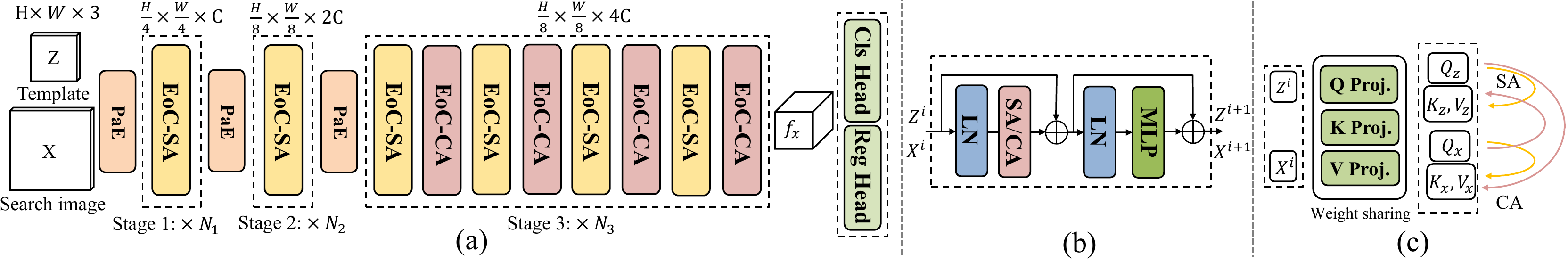}}
		\vspace{-0.8em}
		\caption{(a) architecture of our proposed Single Branch Transformer for tracking. Different from Siamese, DCF and Transformer-based methods, it does not have a standalone module for computing correlation. Instead, it embeds correlation in all Cross-Attention layers which exist at different levels of the networks. The fully fused features of the search image are directly fed to Classification Head (Cls Head) and Regression Head (Reg Head) to obtain localization and size embedding maps. (b) shows the structure of a Extract-or-Correlation (EoC) block. (c) shows the difference of EoC-SA and EoC-CA. PaE denotes patch embedding. LN denotes layer normalization. }
		\label{fig:architecture}
		\vspace{-1.5em}
	\end{figure*}

	The overall framework of SBT is shown in Fig.~\ref{fig:architecture}. It has three model stages on top of Extract-or-Correlation~(EoC) blocks. The patch embedding produces embeddings for the template and search images. Then the embeddings are fed to the stacked EoC blocks.
	There are two variants of EoC, \ie EoC-SA and EoC-CA, which use Self-Attention (SA) and Cross-Attention (CA) as its core operator, respectively.
	The EoC-SA block fuses features within the same image while the EoC-CA block mixes features across images. 
	The output features of the search image are directly fed to the prediction heads to obtain a spatial score map and a size embedding map. 
	Our key technical innovation is introducing one single stream for template and search image pair processing that jointly extract or correlate through homogeneous attention-based blocks. 
	Thus, SBT can be pre-trained on abundant unpaired images such as ImageNet~\cite{ImageNet}, leading to a fast convergence in the fine-tune on tracking.
	
	
	Extensive experiments are conducted to compare different SBT network designs. Based on the insights, we summarize a number of general principals.
	Our method achieves superior performance and improves Siamese, DCF and Transformer-based trackers as can be seen in Fig. \ref{fig:bab}. 
	The main contributions of this work are as follows:
	
	\begin{itemize}[leftmargin=0.4cm]
		\vspace{-0.4em}
		\item We present a novel tracking framework which allows the features of the search and template image to be deeply fused for tracking. It further improves existing popular tracking pipelines. To our best, we are the first to propose a specialized target-dependent feature network for VOT.  
		
		\vspace{-0.4em}
		\item We conduct a systematic study on SBT tracking both experimentally and theoretically, and summarize several general principles for following works.
	\end{itemize}
	
	\vspace{-0.2em}
	The rest of the paper is organized as follows. We discuss related work in Sec.~\ref{sec:related}. The SBT framework is presented in Sec.~\ref{sec:Architecture}. Then, we conduct empirical studies and theoretical analysis on SBT in Sec.~\ref{sec:empirical} and Sec.~\ref{sec:theory}, respectively. Finally, we provide extensive experimental results in Sec.~\ref{sec:experiment} and conclude the paper in Sec.~\ref{sec:conclusion}.

	\section{Related Work}
	\label{sec:related}

	\noindent \textbf{Visual Tracking.}
	The Siamese network~\cite{SiameseFC} based trackers have drawn great attention in recent years. 
	By introducing the powerful backbones~\cite{siamrpn++, siamdw} and elaborated prediction networks~\cite{siamrpn, siamfcpp, SiamCAR}, Siamese trackers obtain superior performance. 
	However, the offline target matching with a shallow correlation structure~\cite{SiameseFC} lacks of discriminative power towards distractors.  
	Then, the dedicated modifications rise, including attention mechanism~\cite{attnsiam, deforsiam, gat}, online module~\cite{drol, Ocean}, cascaded frameworks~\cite{spm, CascadedSiameseRPN, siamrn}, update mechanism~\cite{ UpdateNet} and target-aware model fine-tuning~\cite{ targetaware, MAML-track}.
	Despite the improvements, most of them bring much complexity to the Siamese tracking pipeline.  
	Instead, our target-dependent feature network can upgrade the original network seamlessly. Moreover, our feature network formulates a novel and conceptually simple tracking pipeline by removing the separated correlation step in Siamese trackers.

	Discriminative Correlation Filter (DCF) tracker~\cite{KCF} learns a target model by solving least-squares based regression online. It is further improved by fast gradient algorithm~\cite{atom}, end-to-end learning~\cite{DiMP, DCFST} and CNN-based size estimation~\cite{SuperDiMP, dtt}. 
	However, DCF is highly sensitive to the complex handcrafted optimization, as well as the quality of features which may lack of instance-level discrimination under challenging scenarios. 
	To improve this, our discriminative target-dependent features can greatly lighten the burden for the online DCF. 
	
	Recent rising Transformer-based methods~\cite{tmt, transt, stark, transdcf, dualtfr} exploit the long-range modelling of Transformer to effectively fuse the features. Thus, they can track robustly without online learning.  
	However, the Transformer~\cite{vaswani2017attention} mainly designed for language processing domain is difficult to be initialized properly for vision tasks during training, resulting in enormous costs. 
	Instead of using Transformer as fusion module~\cite{transt, stark, dtt}, we leverage the attention scheme to dynamically generate customized features which establish the hierarchical fine-grained correspondence between target and search area. 
	
	\noindent \textbf{Vision Backbone.} 
	Modern CNNs~\cite{googlenet, ResNet} generally serve as the backbone network in vision tasks.
	Recently, Vision Transformer (ViT)~\cite{vit, swin, pvt}, guided by the principles from CNN, achieves impressive results as vision backbone.
	Deeper and more effective architectures are the two pillars of powerful backbones, which boost numerous downstream tasks. 
	Similarly, the improvements brought by powerful backbone in VOT mainly attribute to the more expressive feature embedding~\cite{siamdw, siamrpn++}, which has subtle differences to other tasks, \eg object detection. 
	However, the dynamic nature of VOT actually requires asymmetrical encoding for template and search image, which has not been given sufficient attention in most prior works.  
	By considering that, we propose a dynamic instance-varying backbone for VOT, beyond only pursuing an expressive embedding.
	
	\section{Architecture}
	\label{sec:Architecture}
	This section introduces the overall architecture of our Single Branch Transformer (SBT) (Fig.~\ref{fig:architecture}) as well as its main building block (EoC block). Then, in the next section, we evaluate a number of instantiations of the architecture followed by a summary of favorable design principals.

	\subsection{Patch Embedding}
	
	\label{sec:Embedding}
	Our model takes two images as input, comprising a template image $z \in \mathbb{R}^{3 \times H_{z} \times W_{z}}$ and a larger search image $x \in \mathbb{R}^{3 \times H_{x} \times W_{x}}$. In general, $z$ is centered on the target object while $x$ represents a larger region in the subsequent frame which contains the target.
	In the Patch Embedding (Pa.E) stage, the two images are fed to a convolutional layer $\varphi_{p}^{0}$ with kernel size $7 \times 7$ and stride $4$, followed by a layer normalization (LN) layer. It embeds the images into feature maps of $f_{z}^{0}$ and $f_{x}^{0}$, respectively.
	\begin{equation}
	f_{z}^{0}, f_{x}^{0} = {\rm LN}(\varphi_{p}^{0}(z)),{\rm LN} (\varphi_{p}^{0}( x)), 
	\label{Eq1}
	\end{equation}
	where $f_{z}^{0} \in \mathbb{R}^{C_{0} \times {\frac{H_z}{4}} \times {\frac{W_z}{4}} }, f_{x}^{0} \in \mathbb{R}^{C_{0} \times {\frac{H_x}{4}} \times {\frac{W_x}{4}} }$ and $C_{0}$ is the number of channels.

	%

	\begin{table*}[t]
		\begin{center}
			\caption{The left part compares different factors of SBT including attention computation methods (ATTN), position encoding methods (PE), patch embedding methods (PaE), number of model parameters and flops. The right part compares the rest of factors based on $\rm A_5$ (described in the left part) such as the feature dimensions (DIM) and the number of blocks (BLK), as well as the stride of the feature maps in each stage. All models unless explained follow the same setting: training from scratch, interleaved EoC-SA/EoC-CA block in the third stage, $128 \times 128 $ for template image and $256 \times 256$ search image.} 
			\vspace{-2mm}
			\resizebox{1.95\columnwidth}{!}{%
				
				\fontsize{8.0 pt}{3.5mm}\selectfont
				\begin{threeparttable}
					\setlength{\tabcolsep}{0.01mm}
					
					\begin{tabular}{ @{}c@{}  
							@{}c@{} @{}c@{} @{}c@{} @{} c@{} 
							@{}c@{} @{}c@{} @{}c@{} @{}c@{} 
							@{}c@{}  @{}c@{} @{}c@{} @{}c@{} 
							@{}c@{} @{}c@{} @{}c@{} @{}c@{} 
							@{}c@{} @{}c@{} @{}c@{} @{}c@{}}
						\toprule[0.08em]
						\textbf{Setting}
						& ~$\rm A_{1}$\tnote{1}~ & ~$\rm A_{2}$\tnote{2}~ & ~$\rm A_{3}$~
						& ~$\rm A_{4}$~ & ~$\rm A_{5}$~  & ~$\rm A_{6}$~
						& ~$\rm A_{7}$~
						& ~~  
						& \textbf{Setting}
						& $\rm B_{1}$ & $\rm B_{2}$ & $\rm B_{3}$  
						& $\rm B_{4}$ & $\rm B_{5}$ &$\rm B_{6}$ 
						& $\rm B_{7}$ & $\rm B_{8}$
						\\
						\midrule[0.06em]
						\textbf{Refer to}
						&~\cite{vit}~
						& ~\cite{swin}~
						& ~\cite{rest}~ 
						&~\cite{pvt}~ &~\cite{pvt}~ &~\cite{pvt}~ &~\cite{twins}~
						& ~~  
						&~\textbf{DIM(1,2)}~
						&\cellcolor{cyan!20}[64, 128] & \cellcolor{cyan!20}[64, 128] & \cellcolor{cyan!20}[64, 128]
						& \cellcolor{cyan!20}[64, 128] & \cellcolor{cyan!20}[64, 128] & \cellcolor{cyan!20}[64, 128]
						& \cellcolor{cyan!20}[64, 128]  
						& [32, 64]
						
						\\
						~\textbf{ATTN}~
						&~VG~ &~SL~ & \cellcolor{cyan!20}~SRG~ &\cellcolor{cyan!20}~SRG~
						&\cellcolor{cyan!20}~SRG~ &\cellcolor{cyan!20}~SRG~ &~VL/SRG~
						
						& ~~  
						
						&~\textbf{DIM(3,4)}~
						&\cellcolor{cyan!20}~[320]~ & ~[320,512]~ & ~[320,512]~
						& ~[512]~ & \cellcolor{cyan!20}~[320]~ &~[320,512]~
						&\cellcolor{cyan!20}~[320]~
						&\cellcolor{cyan!20}~[320]
						
						\\
						\textbf{PE} &~Abs
						& ~Rel~ & \cellcolor{cyan!20}~Cond~
						& \cellcolor{cyan!20}~Cond~ & \cellcolor{cyan!20}~Cond~
						& ~Rel~ & \cellcolor{cyan!20}~Cond~
						& ~~  
						&\textbf{BLK}
						& \cellcolor{cyan!20}[3,4,10] & [4,2,6,1] & [2,2,6,2]
						& [2,2,4] & \cellcolor{cyan!20}[3,4,10] & [2,4,6,1]
						& [3,4,12] 
						& [3,4,10] 
						\\
						\textbf{PaE}
						& ~$H_{1}$\tnote{3}~ & ~$H_{2}$\tnote{3}~ &\cellcolor{cyan!20}~~Conv~ & ~$H_{2}$\tnote{3}~ 
						& \cellcolor{cyan!20}~~Conv~ & \cellcolor{cyan!20}~~Conv~ & \cellcolor{cyan!20}~~Conv~
						& ~~  
						&\textbf{STR}
						& \cellcolor{cyan!20}[4,2,1] & [4,2,1,1] & [4,2,1,1] 
						& \cellcolor{cyan!20}[4,2,1] & [4,1,2] & [4,2,1,1] 
						& [4,2,2]\tnote{4}
						& \cellcolor{cyan!20}[4,2,1] 
						\\
						
						\midrule[0.06em]
						
						~~\textbf{Param.(M)}~ 
						& \cellcolor{red!5}~~22.5~ & ~40.2~ &\cellcolor{red!5} ~23.9~
						&  \cellcolor{red!20}~20.1~ & \cellcolor{red!10}~21.3~ & \cellcolor{red!10} ~21.0~
						&  \cellcolor{red!30}~19.6~
						&~~  
						&~~\textbf{Param.(M)}~ 
						&\cellcolor{red!5}~~21.3~& \cellcolor{red!30}~18.6~ &\cellcolor{red!5}~21.1~
						& \cellcolor{red!10}~20.5~&\cellcolor{red!10}~20.8~ &  \cellcolor{red!20}~19.3~
						& \cellcolor{red!10}~20.8~
						& \cellcolor{red!40}~15.1~
						\\
						~\textbf{Flops(G)}~
						& ~~35.1~ & ~36.5~ & \cellcolor{red!10}~20.2~ 
						&  \cellcolor{red!20}~18.9~ & \cellcolor{red!10}~19.6~ &  \cellcolor{red!10}~19.3~
						&  \cellcolor{red!30}~17.5~
						& ~~  
						&~\textbf{Flops(G)}~
						& \cellcolor{red!20}~~19.6~ &  \cellcolor{red!20}~19.3~ & \cellcolor{red!10}~22.5~
						&  \cellcolor{red!20}~19.2~ & \cellcolor{red!5}~24.4~ & \cellcolor{red!5}~24.7~
						&  \cellcolor{red!30}~12.1~
						& \cellcolor{red!40}~14.5~
						\\
						\textbf{AO}
						& ~~~47.5~ & ~56.4~ & \cellcolor{red!30}~63.7~
						& \cellcolor{red!10}~61.7~ & \cellcolor{red!20}~63.5~ &\cellcolor{red!20}~63.1~
						& \cellcolor{red!5}~60.1~ 
						& ~~    
						& \textbf{AO} 
						&  ~~\cellcolor{red!30}~63.5~ & \cellcolor{red!5}~57.4~ &\cellcolor{red!10}~60.9~ 
						& \cellcolor{red!5}~56.7~ &  \cellcolor{red!30}~63.3~ & \cellcolor{red!10}~60.6~  
						& ~52.2~
						&\cellcolor{red!5} ~56.2~
						\\
						\bottomrule[0.08em]
						
					\end{tabular}
					
					\begin{tablenotes}
						\footnotesize
						\item[1] $\rm A_{1}$ does not have hierarchical structure, so we adopt $4$ downsampling ratio at the beginning and drops the classification token.
						\item[2] For $\rm A_{2}$, we set the same image size $(224 \times 224)$ for template and search image for simplicity.
						\item[3] $H_{1}$ denotes the $\rm A_{1}$ splits an input image into non-overlapping patches ($4 \times 4$). $H_{2}$ denotes a linear layer to change dimensions after patch split.  
						\item[4] For model settings with total network stride 16, we increase the search image size to $320 \times 320$ for a fair comparison.   
					\end{tablenotes}
					
				\end{threeparttable}
			}
			\label{tab:vit}
			
			\vspace{-2.2em}
		\end{center}
		
	\end{table*}

	\subsection{Extract-or-Correlation Block} 
	\label{sec:CASA}
	EoC block which can simultaneously implement Self-Attention (SA) and Cross-Attention~(CA) is the main building block. 
	Intuitively, they gradually fuse features from the same and different images, respectively. 
	It is known that computing attention globally among all tokens leads to quadratic complexity~\cite{swin}. To address this, there are a number of works which attempt to reduce the computation cost. We present a general formulation for different efficient attention methods. On top of the formulation, we describe our SA and CA operations.


	Let $\chi_{(.)}$ denote a function that reshapes/arranges a feature maps into the desired form. The function varies for different methods. We compute the $q, k, v$ features as:
	\begin{equation}
	\begin{aligned}
	& q_{i}= [\chi_{q} (f_{i})] ^\mathsf{T}\omega_{q},\quad i \in \{z, x\}, \\
	& k_{i}= [\chi_{k} (f_{i})] ^\mathsf{T}\omega_{k},\quad i \in \{z, x\}, \\
	& v_{i}= [\chi_{v} (f_{i})] ^\mathsf{T}\omega_{v},\quad i \in \{z, x\}, 
	\end{aligned}
	\end{equation}
	where $\{\omega_{q},\omega_{k},\omega_{v}\}$ represent linear projections. 
	
	The Vanilla Global attention (VG)~\cite{vit} computes attention among all tokens. So $\{\chi_{q},\chi_{k},\chi_{v}\}$ represent identity mapping. The Spatial-Reduction Global attention (SRG)~\cite{pvt, rest} uses a convolution with a stride larger than one (\ie $\{\chi_{k},\chi_{v}\}$) to reduce the spatial resolution of the key and value features. The resolution of the query features is not changed. Then it computes global attention as VG. The method largely reduces the computational overhead. The Vanilla Local window attention (VL)~\cite{twins} splits feature tokens in groups based on their spatial locations and only computes attention within each group.
	Swin Transformer~\cite{swin} further adds a Shift window mechanism to vanilla Local attention~(SL) for global modelling.

	Since the target object may appear anywhere in the search image, it is not practical to use local attention methods for CA. In our work, we use SRG to implement SA and CA. More discussions are in Sec.~\ref{sec:empirical}. The following equation shows how we compute SA or CA: 
	\begin{equation}
	\begin{aligned}
	& \tilde f_{ij}  = {\rm Softmax} (\frac{q_{i} k_{j}^\mathsf{T} }{\sqrt{d_{h}}}) v_{j}, \quad i, j \in \{z, x\}, \\
	\end{aligned}
	\end{equation}
	In SA, $i$ and $j$ are from the same source (either $z$ or $x$) and the resulting feature update is:
	\begin{equation}
	\begin{aligned}
	& f_{z} :=  f_{z} + \tilde f_{zz},\quad f_{x} :=  f_{x} + \tilde f_{xx}, 
	\end{aligned}
	\end{equation}
	In CA, it mixes the features from different sources: 
	\begin{equation}
	\begin{aligned}
	& f_{z} :=  f_{z} + \tilde f_{zx},\quad f_{x} :=  f_{x} + \tilde f_{xz}.
	\end{aligned}
	\end{equation}

	We can see that the correlation between the two images is deeply embedded in feature extraction seamlessly. EoC block also consists of two LN layers and a 2-layer MLP as shown in Fig~\ref{fig:architecture} (b). 
	
	\subsection{Position Encoding}
	%
	%
	For majority methods~\cite{vit, swin, detr}, the encoding is generated by the sinusoidal functions with Absolute coordinates~(Abs) or Relative distances~(Rel) between tokens.
	Being much simpler, Conditional positional encoding~\cite{condPE, pvt, rest}~(Cond) generates dynamic encoding by convolutional layers.
	In our model, we add a $3 \times 3$ depth-wise convolutional layer $\varphi_{pe}$ to MLP before GELU as conditional PE.

	\subsection{Direct Prediction}
	Different from the existing tracking methods, we directly add a classification head $\Phi_{cls}$ and regression head $\Phi_{reg}$ on top of the search feature $\hat f_x$ from SBT $\Omega$ without additional correlation operations:
	\begin{equation}
	\begin{aligned}
	& \hat f_x= \Omega(z, x),  \quad y_{reg} = \Phi_{reg}(\hat f_x), \quad y_{cls} =\Phi_{cls}(\hat f_x), 
	\end{aligned}
	\end{equation}
	where $y_{reg}, y_{cls}$ denote the target regression and classification results to estimate the location and shape of the target.

	We implement $\Phi_{reg}$ and $\Phi_{cls}$ by stacking multiple Mix-MLP Blocks~(MMB) which can jointly model the dependency between the spatial and channel dimensions of the input features $\hat{f}^{i-1}$ in the $i^{th}$ MMB:
	\begin{equation}
	\begin{aligned}
	&\hat{f}^{i}=  \varphi_{sp} ({\rm RS}( \varphi_{cn} ({\rm RS}(\hat{f}^{i-1})))), \\
	\end{aligned}
	\end{equation}
	where $\varphi_{sp}$ and $\varphi_{cn}$ consist of a linear layer followed by RELU activation. $\rm RS$ represents reshape. $\varphi_{cn}$ is applied to features along the channel dimension, and the weights are shared for all spatial locations. In contrast, the operator $\varphi_{sp}$ is shared for all channels.

	\begin{figure*}[t]
		\centering{\includegraphics[scale = 0.51]{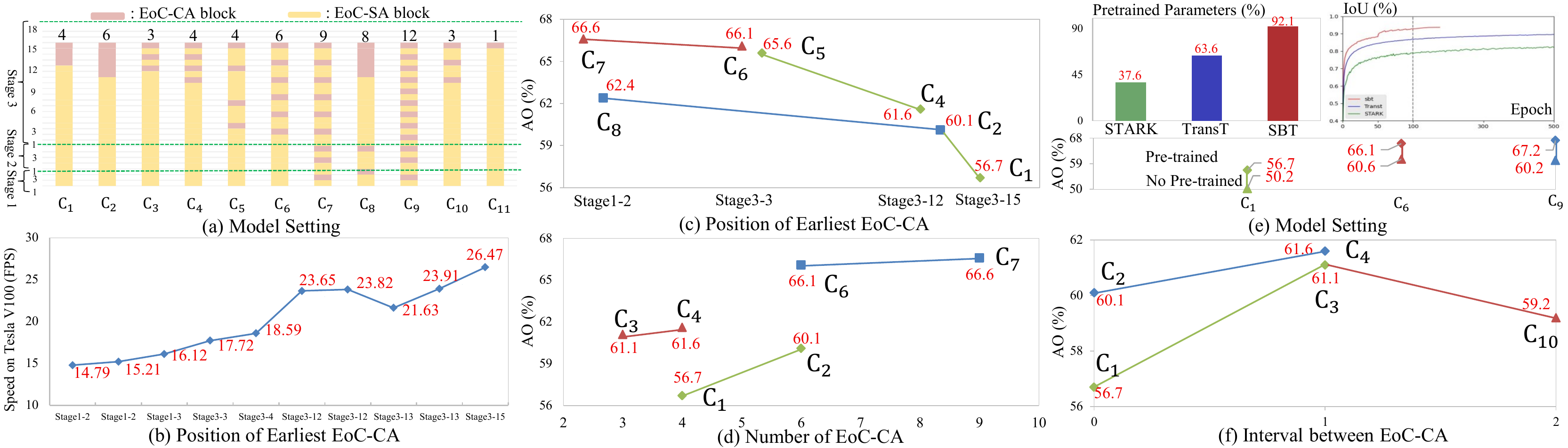}}
		\vspace{-2.0mm}
		\caption{Studies on the number/position of EoC-CA block. (a): different model settings, (b) speed Vs. different model settings,(c): tracking performance Vs. position of earliest EoC-CA block, (d): tracking performance Vs. number of EoC-CA block, (e): tracking performance Vs. pre-trained or not, (f): tracking performance Vs. intervals between EoC-CA block.}
		\vspace{-5mm}
		\label{fig:main}
	\end{figure*}

	\section{Empirical Study of SBT Instantiations}
	\label{sec:empirical}
	
	In this section\footnote{All the experiments follow the official GOT-10k~\cite{GOT10K} test protocol.}, we conduct empirical studies on SBT variants by raising a number of questions. 
	
	
	%
	
	As efficient attention computing is vital to the SBT, 
	we firstly ablate other network factors including hierarchical structure, position encoding and patch embedding.
	As shown in Tab.~\ref{tab:vit},
	it is obvious that \emph{hierarchical structure performs much better than singe stage because of multi-scale representation} ($\rm A_{1}$ Vs. $\rm A_{2}$ to $\rm A_{7}$).
	Conditional PE only surpasses the relative PE by $0.4$ points ($\rm A_{5}$ Vs. $\rm A_{6}$). 
	The difference between PE methods is rather small, indicating that \emph{PE does not have key impacts on performance}.   
	We also find that \emph{convolutional PaE is more practical and expressive than hand-crafted patch merging}~($\rm A_{4}$ Vs. $\rm A_{5}$).

	\textbf{Which attention computation is better for SBT tracker?} 
	The main difference between attention computation lies in the operation to reduce complexity~(global/local attention). 
	We find that local attention (VL/SL) block cannot directly perform Cross-Attention as the inequality of local windows in template and search image. 
	Thus, for SBT constructed by pure local attention blocks, we adopt same image size ($224 \times 224$) for template/search image ($\rm A_{2}$) to avoid tedious hand-crafted cross strategies.
	Comparing to the settings with global attention block (VG/SRG) ($\rm A_{3}$ to $\rm A_{7}$) which adopt $128\times 128$ as template size, the performance of pure local attention ($\rm A_{2}$) drop at least $3.6$ points in $AO$ with more parameters and flops. 
	\emph{This is mainly due to the negative impacts of over background information in template which may confuse the search branch.} 
	We also investigate the mix setting of SRG and VL block ($\rm A_{7}$). 
	To be specific, the VL block is for Self-Attention while SRG block is for Cross-Attention.
	We observe that the pure SRG block design achieves the better performance ($\rm A_{5}$ Vs. $\rm A_{7}$).
	This illustrates that \emph{SBT benefits from unified block choice.}
	$\rm A_{3}$ also validates the effectiveness of pure SRG blocks with $63.7\%$ in $AO$.
	We conclude that \emph{pure SRG block is more practical and efficient for SBT tracker.}
	
	\textbf{Do earlier and more EoC-CA blocks help to tracking better?}
	With a baseline designed from above principles, it strikes to us that SBT may benefit from earlier and more cross correlation. 
	We ablate different position/number of EoC-CA block in Fig.~\ref{fig:main}.
	As shown in Fig.~\ref{fig:main} (d), when the number of EoC-CA blocks increases, the performance of model rises consistently with the same EoC-SA/EoC-CA position pattern ($\rm C_{3}$ vs. $\rm C_{4}$, $\rm C_{1}$ vs. $\rm C_{2}$, $\rm C_{6}$ vs. $\rm C_{9}$). 
	It proves that \emph{SBT tracker benefits from more comprehensive Cross-Attention between template and search branch.}
	In Fig.~\ref{fig:main} (d), when the number of EoC-CA block is the same, earlier cross design has significant positive impacts ($\rm C_{4}$ surpasses $\rm C_{1}$ by $4.9$ points, $\rm C_{6}$ surpasses $\rm C_{2}$ by $6.5$ points).
	The underlying reason is that \emph{early-cross generates target-dependent features which help tracker to see better.}

	\textbf{Is tracking performance related to the placement pattern of EoC-CA blocks?} 
	As the position and number of EoC-CA block have significant impacts on performance, it comes to us which pattern of placement is the optimal choice. 
	So we make attempts to place EoC-SA/EoC-CA block differently. 
	In Fig.~\ref{fig:main} (f), we surprisingly find that interleaved EoC-SA/EoC-CA pipeline performs better than the separation pattern even with less Cross-Attention and latter earliest cross position ($\rm C_{3}$ vs. $\rm C_{1}$).
	The potential cause is that \emph{EoC-SA block can refine the template/search feature after the correlation, resulting in a more expressive feature space for matching.} 
	In Fig.~\ref{fig:main} (f), model ($\rm C_{9}$) achieves the best performance $67.2\%$ when the interval is $1$. 
	When the interval increases to 2, the performances drops from  $61.1\%$ to $59.2\%$ ($\rm C_{3}$ vs. $\rm C_{10}$).
	Thus, \emph{we prefer an interleaved EoC-SA/EoC-CA block design for SBT tracker.}   
	
	\textbf{What is the optimal network variants for tracking model?} 
	Then, it comes to us a long-standing problem for designing a deep tracker.
	We ablate different network stride, model stage and model size. 
	As shown in Tab.~\ref{tab:vit}, \emph{over parameters and flops in shallow level (stage 1 and 2) is harmful.}
	It is mainly because the low dimension cannot formulate informative representations ($57.4$ of $\rm B_{2}$ Vs. $60.6$ of $\rm B_{6}$). 
	We also observe that increasing the head number slightly improves the performance but decreases the speed.
	\emph{With the same total network stride, three-stage model performs better than four-stage model} ($63.5$ of $\rm B_{1}$ Vs. $57.4$ of $\rm B_{2}$) with comparable parameters and flops. 
	Though setting the network stride to 16 can reduce the flops, the performance drops $11.3$ points ($\rm B_{1}$ Vs. $\rm B_{7}$), indicating that \emph{SBT tracker prefers larger spatial size of features.}
	As the channel dimensions influence model size a lot, \emph{it is vital to achieve a balance between block numbers and channel dimensions} ($56.7$ of $\rm B_{4}$ Vs. $63.3$ of $\rm B_{5}$).
	
	\textbf{Does flexible design of EoC-SA/EoC-CA bring negative/positive effects?}
	We examine the potential negative/positive effects in SBT. 
	From Fig.~\ref{fig:main} (b) and Fig.~\ref{fig:main} (c), we observe that early-cross in shallow-level (stage 1 and 2) does not bring much improvements ($\rm C_{2}$ vs. $\rm C_{8}$, $\rm C_{6}$ vs. $\rm C_{7}$) but lowers the inference speed. 
	It is because the early-cross destroys the one-shot inference. 
	The shallow-level EoC-SA blocks perform as buffer. 
	\emph{So the trade-off between early-cross and speed should be well-considered.}
	In Fig.~\ref{fig:main} (e),
	\emph{SBT tracker benefits from more pre-trained weights and converges much faster than Transformer-based trackers,} such as TransT~\cite{transt} and STARK~\cite{stark}.  

	\section{Single Branch Transformer Driven Tracking}
	\label{sec:theory}
	Beyond exploring SBT experimentally, we theoretically analyze SBT from the perspective of general VOT. Then, we design four versions of SBT and integrate them into typical trackers to show generality. 
	
	\subsection{Theoretical Analysis on SBT for Tracking}
	\textbf{SBT overcomes the intrinsic restrictions in deep trackers.}
	Deep trackers have an intrinsic requirement for strict translation invariance, 
	$f\left(c, x\left[\Delta \tau_{j}\right]\right)=f(c, x)\left[\Delta \tau_{j}\right]$
	where $\Delta \tau_{j}$ is the shift window operator, $c$ denotes the template/online filter in Siamese/DCF tracking and $f$ denotes correlation operation. 
	Modern backbones~\cite{ResNet, resnext} can provide more expressive features while their padding is inevitable to destroy the translation invariance.
	Thus, deep trackers~\cite{siamdw, siamrpn++} crop out padding-affected features and adopt spatial-aware sampling training strategy to keep the translation invariance.
	Theoretically, padding in SBT can be removed completely or only exists in patch embedding for easy implementation. 
	Moreover, the flattened feature tokens has permutation invariance which makes EoC block completely translation invariance. 
	As the EoC block provides global receptive field, SBT can enjoy arbitrary size of template/search image and larger search area scale.
	Thus, we argue that SBT driven tracking can overcome the intrinsic restrictions in classical deep trackers theoretically by using brand-new network modules.

	\textbf{Cross-Attention is more than twice as effective as depth-wise correlation.}
	We first prove that Cross-Attention can be decomposed into dynamic convolutions (D-Conv).
	CA which performs as feature correlation is mathematically equivalent to two D-Convs and a SoftMax layer. 
	For simplicity, we annotate the encoded $\{q,k,v\}$ features to their original feature as the projection matrix is $1 \times 1$ position-wise convolutional filters. 
	So the CA for query from search feature $x$ to template feature $z$ is:
	\begin{equation}
	\begin{aligned}
	&{\rm Inter} = {{\rm RS} (z)}^Tx+\bm{0} = {\bm W_{1}}(z)^T x+\bm{b_{1}}, \\
	&{\rm Attn}_{xz} = {\rm Softmax} ({{\rm Inter}}),\\
	& \tilde f_{xz} = {{\rm Attn_{xz}} z} + x = \bm{W_{2}}(z, x)^Tx+{\bm b_{2}}(x),
	\end{aligned}
	\label{early}
	\end{equation} 
	where $\bm{W}(a, b), \bm{b}(a, b)$ is the weight matrix and bias vector of dynamic filters generated by $\{a, b\}$ and $\rm RS$ denotes reshape. 
	To obtain the correlation feature $\tilde f_{xz}$, the search feature $x$ goes through a D-Conv layer generated by $z$, a SoftMax layer and another D-Conv layer generated by $z$ and $x$. Two D-Conv layers come from the reshape of $\rm z$ along channel and spatial dimension. 
	The depth-wise correlation (DW-Corr) or pixel-wise correlation (Pix-Corr)~\cite{AlphaRefine} is only equivalent to one D-Conv layer.
	Thus, CA is twice as effective as DW-Corr or Pix-Corr with the same template feature as dynamic parameters. 
	
	\textbf{Hierarchical feature utilization is embedded in serial pipeline.}
	Siamese trackers~\cite{siamrpn++, siamban} perform correlation for each hand-selected feature pair and feed them into parallel prediction heads. Then, prediction results are aggregated by a weighted sum.    
	Comparing to the hand-craft layer-wise aggregation, SBT structure explores multi-level feature correlation intrinsically. We take three-level feature utilization as an example: 
	\begin{equation}
	\begin{aligned}
	&{x}_{i}, {z}_{i}= \phi_{ca}^{i}({\tilde {x}_{i}}, {\tilde {z}_{i}}), \quad i \in \{0, 1, 2\}\\
	&{x}_{2}, {z}_{2} = \phi_{ca}^{2} ( \phi_{ca}^{1} ( \phi_{ca}^{0} ( {x}_{0}, {z}_{0} ))),\\
	&{S}_{sbt}= \varphi^{p} ({{x}_{2}}),
	\end{aligned}
	\label{early}
	\end{equation} 
	where $\{0, 1, 2\}$ represents shallow, intermediate and deep level, $\{\tilde {x}, \tilde {z}\}$ are the previous layer features of $\{{x},  {z}\}$, $\{\phi_{ca}, \varphi^{p}\}$ denote EoC-CA block and prediction head. 
	Though in a serial pipeline, the prediction result ${S}_{sbt}$ naturally contains hierarchical feature correlation results. 
	%

	\begin{table}[t]
		\begin{center}
			\fontsize{5.0pt}{2.5mm}\selectfont
			\setlength{\tabcolsep}{0.01mm}
			\resizebox{0.98\columnwidth}{!}{%
				
				\begin{tabular}{@{}c@{}|@{}c@{}|@{}c@{}|@{}c@{}|@{}c@{}@{}c@{}@{}c@{}}
					\cline{1-5}
					~~~~
					& \multicolumn{1}{@{}c@{}|}{~~~Light~~~} 
					& \multicolumn{1}{@{}c@{}|}{~~~Small~~~} 
					&  \multicolumn{1}{@{}c@{}|}{~~~Base~~~} 
					&  \multicolumn{1}{@{}c@{}}{~~~Large~~~}
					&
					\\ \cline{1-5}
					{\multirow{1}{*}{PaE}} 
					& ${\rm Conv}(7, 32, 4)$
					
					& ${\rm Conv}(7, 64, 4)$
					& ${\rm Conv}(7, 64, 4)$
					& ${\rm Conv}(7, 64, 4)$
					
					\\ \cline{1-5}
					{\multirow{1}{*}{Stage1}} 
					& $\begin{bmatrix}
					\!\text{EoCA\_1\_8}\! \\ 
					\!\text{MLP\_32}\!  \\
					\end{bmatrix} \times 2$ ~~
					
					& $\begin{bmatrix}
					\!\text{EoCA\_1\_8}\! \\ 
					\!\text{MLP\_64}\!  \\
					\end{bmatrix} \times 2$ ~~
					
					& $\begin{bmatrix}
					\!\text{EoCA\_1\_8}\! \\ 
					\!\text{MLP\_64}\!  \\
					\end{bmatrix} \times 3$ ~~
					
					& $\begin{bmatrix}
					\!\text{EoCA\_1\_8}\! \\ 
					\!\text{MLP\_64}\!  \\
					\end{bmatrix} \times 3$ ~~
					
					\\ \cline{1-5}
					{\multirow{1}{*}{PaE}} 
					& ${\rm Conv}(3, 64, 2)$
					& ${\rm Conv}(3, 128, 2)$
					& ${\rm Conv}(3, 128, 2)$
					& ${\rm Conv}(3, 128, 2)$
					\\ \cline{1-5}
					{\multirow{1}{*}{Stage2}} 
					
					& $\begin{bmatrix}
					\!\text{EoCA\_2\_4}\! \\ 
					\!\text{MLP\_64}\!  \\
					\end{bmatrix} \times 2$ ~~
					
					& $\begin{bmatrix}
					\!\text{EoCA\_2\_4}\! \\ 
					\!\text{MLP\_128}\!  \\
					\end{bmatrix} \times 2$ ~~
					
					& $\begin{bmatrix}
					\!\text{EoCA\_2\_4}\! \\ 
					\!\text{MLP\_128}\!  \\
					\end{bmatrix} \times 4$ ~~
					
					& $\begin{bmatrix}
					\!\text{EoCA\_2\_4}\! \\ 
					\!\text{MLP\_128}\!  \\
					\end{bmatrix} \times 4$ ~~
					&
					\\ \cline{1-5}
					{\multirow{1}{*}{PaE}}
					& ${\rm Conv}(3, 160, 1)$
					& ${\rm Conv}(3, 320, 1)$
					& ${\rm Conv}(3, 320, 1)$
					& ${\rm Conv}(3, 320, 1)$
					\\ \cline{1-5}
					{\multirow{1}{*}{Stage3}} 
					
					& $\begin{bmatrix}
					\!\text{EoCA\_5\_2}\! \\ 
					\!\text{MLP\_160}\!  \\
					\end{bmatrix} \times 6 $~~

					& $\begin{bmatrix}
					\!\text{EoCA\_5\_2}\! \\ 
					\!\text{MLP\_320}\!  \\
					\end{bmatrix} \times 6$ ~~
					
					& $\begin{bmatrix}
					\!\text{EoCA\_5\_2}\! \\ 
					\!\text{MLP\_320}\!  \\
					\end{bmatrix} \times 10$ ~~
					
					& $\begin{bmatrix}
					\!\text{EoCA\_5\_2}\! \\ 
					\!\text{MLP\_320}\!  \\
					\end{bmatrix} \times 18$ ~~
					
					\\ \cline{1-5}
					{\multirow{1}{*}{PaE}}
					& ${\rm Conv}(3, 256, 2)$
					& ${\rm Conv}(3, 512, 2)$
					& ${\rm Conv}(3, 512, 2)$
					& ${\rm Conv}(3, 512, 2)$
					\\ \cline{1-5}
					{\multirow{1}{*}{stage4}} 
					
					& $\begin{bmatrix}
					\!\text{EoCA\_8\_1}\! \\ 
					\!\text{MLP\_256}\!  \\
					\end{bmatrix} \times 2$ ~~

					& $\begin{bmatrix}
					\!\text{EoCA\_8\_1}\! \\ 
					\!\text{MLP\_512}\!  \\
					\end{bmatrix} \times 2$ ~~
					
					& $\begin{bmatrix}
					\!\text{EoCA\_8\_1}\! \\ 
					\!\text{MLP\_512}\!  \\
					\end{bmatrix} \times 2$ ~~
					
					& $\begin{bmatrix}
					\!\text{EoCA\_8\_1}\! \\ 
					\!\text{MLP\_512}\!  \\
					\end{bmatrix} \times 2$ ~~
					
					\\  \cline{1-5}

					{\multirow{1}{*}{Head}} 
					&
					\multicolumn{5}{@{}c@{}}{
						~Classification:~~$\text{MMB} \times 2$ ~~~~~~ Regression:~~$\text{MMB} \times 2$~ 
					}
					\\\cline{1-5}
					EoC-CA~  &[2, 4, 6] &[2, 4, 6] & [2, 4, 6, 8, 10] & ~[6, 8, 10, 12, 14, 16, 18]
					\\\cline{1-5}
					\textbf{Params}~ &3.03 M &13.80 M & 21.27 M& 35.20 M
					\\\cline{1-5}
					\textbf{FLOPs}~  & 3.81 G & 11.92 G& 19.27 G& 31.46 G
					\\\cline{1-5}
					\textbf{Speed}~  & 62 FPS & 50 FPS& 37 FPS & 24 FPS
					\\\cline{1-5}
					
				\end{tabular}
			}
		\end{center}
		
		\vspace{-1.5em}
		\caption{ Model settings of SBT in four scales. ``${\rm Conv}(k,c,s)$" means convolution layers with kernel size $k$, output channel $c$ and stride $s$. ``${\rm MLP}\_c$" is the MLP with hidden channel $4c$ and output channel $c$. ``${\rm EoCA}\_n\_r$" is the EoC attention computation with the number of heads $n$ and down-sampling ratio $r$. EoC-CA blocks are in the third stage. We report the speed in single Tesla V100 GPU.}
		\label{version}
		\vspace{-1.0em}
	\end{table}

	\subsection{Four Versions of SBT network}
	Following the guidelines from Sec.~\ref{sec:empirical}, our four versions of SBT is described in Tab.~\ref{version}.
	For pre-training, we add extra fourth model stage and modify the network stride as~\cite{swin}.
	For fine-tune on tracking, we only use three-stage model and replace the prediction head. 
	%
	
	
	\subsection{Correlation-Aware Feature for Other Trackers}
	We replace the backbone in four typical trackers with SBT, which are named as \textbf{Correlation-Aware Trackers} (CAT). The four trackers are: \textbf{SiamFCpp-CA}. \textbf{DiMP-CA}. \textbf{STARK-CA}. \textbf{STM-CA}.  
	Details are in supplement. 
	

	\begin{table}[t]
		\centering
		\setlength{\tabcolsep}{1mm}
		
		\newcommand{\best}[1]{\textbf{\textcolor{red}{#1}}}
		\newcommand{\scnd}[1]{\textbf{\textcolor{blue}{#1}}}
		\newcommand{\trd}[1]{\textbf{\textcolor{green}{#1}}}
		\newcommand{\opt}[1]{\textbf{\textcolor{violet}{#1}}}
		\newcommand{\fast}[1]{\textbf{\textcolor{orange}{#1}}}
		\newcommand{\dist}{\hspace{4pt}}%
		\setlength{\tabcolsep}{0.5mm}
		
		\resizebox{1.00\columnwidth}{!}{%
			\begin{tabular}{l@{\dist}c@{\dist}c@{\dist}c@{\dist}c@{\dist}c@{\dist}c@{\dist}c@{\dist}c@{\dist}c@{\dist}c@{\dist}c@{\dist}c@{\dist}c@{\dist}c@{\dist}c@{\dist}c@{\dist}c@{\dist}c@{\dist}c@{\dist}c@{\dist}c@{\dist}c@{\dist}c@{\dist}c@{\dist}c@{\dist}c@{\dist}c@{\dist}c@{\dist}c@{\dist}}
				\toprule
				& 
				&  &  
				&  &  
				& Tr  &  
				& Stark  & Stark
				&\cellcolor{gray!20} SBT &\cellcolor{gray!20} SBT 
				& \cellcolor{gray!20}SBT &\cellcolor{gray!20} SBT
				\\
				& SiamRPN++ 
				& ATOM  & DiMP 
				& SAMN & AutoMatch
				& Siam & TransT 
				& s50 & st101 
				&\cellcolor{gray!20} light &\cellcolor{gray!20} small
				&\cellcolor{gray!20} base &\cellcolor{gray!20} large
				\\
				&\cite{siamrpn++} 
				& \cite{atom} & \cite{DiMP}
				&\cite{samn} & \cite{automatch}
				&\cite{tmt} & \cite{transt}
				&\cite{stark} & \cite{stark}
				& & 
				&&     \\
				\midrule
				
				$AO \uparrow\ $       
				& 51.8  
				& 55.6 & 61.1 
				& 61.5 & 65.2 
				& 66.0 & 67.1 
				& 67.2 & \trd{68.8}
				&60.2 & 66.8 
				& \scnd{69.9} & \best{70.4}
				\\
				
				$SR_{50} \uparrow\ $     
				& 61.6   
				& 63.4 & 71.7
				& 69.7 & 76.6
				& 76.6 & 76.8
				& 76.1 &\trd{78.1} 
				& 68.5 & 77.3
				& \scnd{80.4} & \best{80.8}
				\\
				
				$SR_{75} \uparrow\ $     
				& 32.5 
				& 40.2 & 49.2 
				& 52.2 & 54.3 
				& 57.1 & 60.9 
				& 61.2 & \scnd{64.1}
				& 53.0 &59.2 
				& \trd{63.6} & \best{64.7}
				\\\bottomrule
				
			\end{tabular}
		}
		\vspace{-3mm}
		\caption{Comparison on the GOT-10k~\cite{got} test set. 
		}
		\vspace{-3mm}
		\label{tab:got10k}%
	\end{table}
	
	\begin{table}[t]
		\centering
		\setlength{\tabcolsep}{0.1mm}
		
		\newcommand{\best}[1]{\textbf{\textcolor{red}{#1}}}
		\newcommand{\scnd}[1]{\textbf{\textcolor{blue}{#1}}}
		\newcommand{\trd}[1]{\textbf{\textcolor{green}{#1}}}
		\newcommand{\opt}[1]{\textbf{\textcolor{violet}{#1}}}
		\newcommand{\fast}[1]{\textbf{\textcolor{orange}{#1}}}
		\newcommand{\dist}{\hspace{4pt}}%
		\resizebox{1.00\columnwidth}{!}{%
			\begin{tabular}{l@{\dist}c@{\dist}c@{\dist}c@{\dist}c@{\dist}c@{\dist}c@{\dist}c@{\dist}c@{\dist}c@{\dist}c@{\dist}c@{\dist}c@{\dist}c@{\dist}c@{\dist}c@{\dist}c@{\dist}c@{\dist}c@{\dist}c@{\dist}c@{\dist}c@{\dist}c@{\dist}c@{\dist}c@{\dist}c@{\dist}c@{\dist}c@{\dist}c@{\dist}c@{\dist}}
				\toprule
				& 
				&  &  
				&  &  
				& Tr  &  
				&   & Stark
				& \cellcolor{gray!20}SBT &\cellcolor{gray!20} SBT 
				& \cellcolor{gray!20}SBT &\cellcolor{gray!20} SBT
				\\
				& SiamRPN++ 
				& ATOM  & DiMP 
				& AutoMatch & DualTFR
				& Siam & TransT 
				& DTT & s50
				& \cellcolor{gray!20}light & \cellcolor{gray!20}small
				& \cellcolor{gray!20}base & \cellcolor{gray!20} large
				\\
				&\cite{siamrpn++} 
				& \cite{atom} & \cite{DiMP}
				&\cite{automatch} & \cite{dualtfr}
				&\cite{tmt} & \cite{transt}
				&\cite{dtt} & \cite{stark}
				& & 
				& &    \\
				\midrule
				
				AUC $\uparrow\ $   
				& 49.6 
				& 51.5 & 56.9
				& 58.3 & 63.5
				& 62.4 & 64.9
				& 60.1  & \trd{65.8}
				&56.5 &61.1
				&\scnd{65.9} &\best{66.7}
				\\
				Prec$\uparrow\ $      
				& 49.1
				& 50.5 & 56.7
				& 59.9 & 66.5
				& 60.0 & 69.0 
				& - & \trd{69.7}
				& 57.1 & 63.8
				& \scnd{70.0} & \best{71.1}
				\\\bottomrule
				
			\end{tabular}
		}
		\vspace{-3mm}
		\caption{Comparison on the LaSOT~\cite{LaSOT} test set. 
		}
		\vspace{-3mm}
		\label{tab:lasot}%
	\end{table}
	
	\begin{table}[t]
		\centering
		\setlength{\tabcolsep}{0.1mm}
		
		\newcommand{\best}[1]{\textbf{\textcolor{red}{#1}}}
		\newcommand{\scnd}[1]{\textbf{\textcolor{blue}{#1}}}
		\newcommand{\trd}[1]{\textbf{\textcolor{green}{#1}}}
		\newcommand{\opt}[1]{\textbf{\textcolor{violet}{#1}}}
		\newcommand{\fast}[1]{\textbf{\textcolor{orange}{#1}}}
		\newcommand{\dist}{\hspace{4pt}}%
		\resizebox{1.00\columnwidth}{!}{%
			\begin{tabular}{l@{\dist}c@{\dist}c@{\dist}c@{\dist}c@{\dist}c@{\dist}c@{\dist}c@{\dist}c@{\dist}c@{\dist}c@{\dist}c@{\dist}c@{\dist}c@{\dist}c@{\dist}c@{\dist}c@{\dist}c@{\dist}c@{\dist}c@{\dist}c@{\dist}c@{\dist}c@{\dist}c@{\dist}c@{\dist}c@{\dist}c@{\dist}c@{\dist}c@{\dist}c@{\dist}}
				\toprule	   
				& 
				&  &  
				&  
				&   &  
				& 
				& Stark  & Stark
				&\cellcolor{gray!20} SBT &\cellcolor{gray!20} SBT 
				& \cellcolor{gray!20}SBT &\cellcolor{gray!20} SBT
				\\
				& Ocean
				& ATOM  & SiamMask
				& SuperDiMP
				& STM  & DET50 
				& AlphaRef
				& s50 & st101 
				&\cellcolor{gray!20} light &\cellcolor{gray!20} small
				& \cellcolor{gray!20} base & \cellcolor{gray!20} large
				\\
				&\cite{Ocean} 
				& \cite{atom} & \cite{SiamMask}
				& \cite{SuperDiMP}
				&\cite{STM} & \cite{vot2020}
				& \cite{AlphaRefine}
				&\cite{stark} & \cite{stark}
				& & 
				& &  \\
				\midrule
				
				Acc.$\uparrow\ $     
				& 0.693    
				& 0.462 & 0.624
				& 0.492
				& 0.751  & 0.679
				& 0.754
				& \scnd{0.761} & \best{0.763}
				& 0.742 & 0.750
				& 0.752 &\trd{0.753}
				\\
				
				Rob.$\uparrow\ $ 
				& 0.754   
				& 0.734  & 0.648
				& 0.745
				& 0.574 & 0.787
				& 0.777
				& 0.749 & \trd{0.789}
				& 0.712 & 0.775
				& \scnd{0.825} & \best{0.834}
				\\
				
				EAO$\uparrow\ $ 
				&  0.430  
				& 0.271 & 0.321
				& 0.305
				& 0.308 & 0.441
				& 0.482
				& 0.462 & \trd{0.497}
				& 0.415 & 0.477
				& \scnd{0.515} & \best{0.529}
				\\
				\bottomrule
				
			\end{tabular}
		}
		\vspace{-3mm}
		\caption{Results on VOT2020. We use AlphaRefine~\cite{AlphaRefine} to generate mask for VOT benchmark.
		}
		\vspace{-3mm}
		\label{tab:VOT2020}%
	\end{table}
	
	\begin{figure}[t]
		\centering{\includegraphics[scale = 0.35]{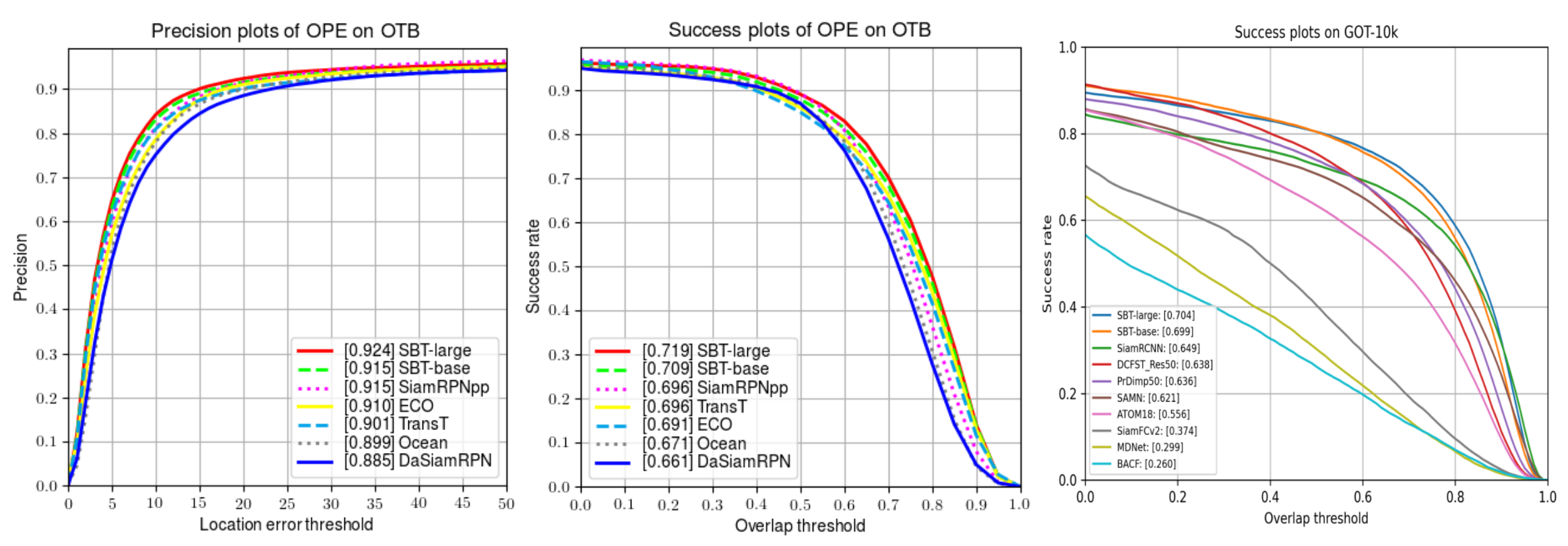}}
		\vspace{-6mm}
		\caption{Comparisons on (a) OTB-100~\cite{OTB2015} and (b) GOT-10k~\cite{GOT10K} test set in terms of success plot.}
		\vspace{-5mm}
		\label{fig:otb}
	\end{figure}

	\section{Experiments}
	\label{sec:experiment}
	
	This section describes the implement details, comparisons to the state-of-the-art (sota) trackers and improvements in CATs.
	Exploration studies are also provided.

	\subsection{Implementation Details} \label{sec:Implementation}
	
	\paragraph{ImageNet pre-training.}
	We firstly train 4-stage SBT with classification head on the ImageNet~\cite{ImageNet}.
	Similar to the network for image classification, our model structure and data flow is one-stream. 
	The setting mostly follows~\cite{imagetrain} and ~\cite{pvt}. We employ the AdamW ~\cite{AdamW} optimizer for $300$ epochs. 
	The input image is resized to $224 \times 224$ and the augmentation and regularization strategies of~\cite{imagetrain} are adopted. 

	\vspace{-3mm}
	\paragraph{Finetune on tracking.}
	Next, the pre-trained weights are to initialize our tracking model. 
	By arranging EoC-SA/EoC-CA blocks, the model is still one-stream in structure but two-stream in data flow.  
	For each image pair, 
	We compute standard cross-entropy loss for the classification and GIoU~\cite{GIoULoss} loss and $L_1$ loss for the regression. 
	We use $8$ tesla V100 GPUs and set the batch size to be $160$. 
	The template and search image size are set to $128 \times 128$ and $256 \times 256$.
	The sample pairs of each epoch is $50, 000$ and the total epoch is $600$. 
	The learning rate is set to be $10^{-4}$ for the head, and $10^{-5}$ for the rest and it decays by a factor of $10$ at the $200^{th}, 400^{th}$ epoch. 
	The training datasets include the train subsets of LaSOT~\cite{LaSOT}, GOT-10K~\cite{GOT10K}, COCO2017~\cite{COCO}, and TrackingNet~\cite{trackingnet}.  
	Other settings are the same with~\cite{dualtfr, transt}. 
	Details are in supplement.

	\begin{table}[t]
		\centering
		\begin{center}
			\setlength{\tabcolsep}{0.01mm}
			\fontsize{8.0pt}{1.10mm}\selectfont
			\newcommand{\dist}{\hspace{0.0pt}}%
			
			\resizebox{1\columnwidth}{!}{
				\begin{threeparttable}
					\begin{tabular}{ @{\dist}c@{\dist} @{\dist}c@{\dist} @{\dist}c@{\dist} @{\dist}c@{\dist} @{\dist}c@{\dist} @{\dist}c@{\dist} @{\dist}c@{\dist} @{\dist}c@{\dist} @{\dist}c@{\dist}  }
						\toprule[0.05em]
						~\textbf{Box-level Tracker}~ &~~~~\textbf{Params(M)}~&~~~\textbf{Flops(G)}~ &~~~$\rm SR_{50}$~&~~~~$\rm SR_{75}$~ &~{AO}~~~
						\\
						\midrule[0.01em]
						~SiamFCpp~
						& ~13.9~
						&~19.8~
						&~~69.5~
						&~47.9~
						&~59.5~
						\\
						\cellcolor{gray!20} ~SiamFCpp-CA~
						& \cellcolor{gray!20}~16.3~
						& \cellcolor{gray!20}~14.1\color{blue}{(5.7$\downarrow$)}~
						& \cellcolor{gray!20}~74.8\color{red}{(5.3$\uparrow$)}~
						& \cellcolor{gray!20}~54.5\color{red}{(6.6$\uparrow$)}~
						& \cellcolor{gray!20}~64.7\color{red}{(5.2$\uparrow$)}~
						\\
						~DiMP~ 
						& ~~26.1~
						& ~-~
						&~71.7~
						&~49.2~
						&~61.1~
						\\
						
						\cellcolor{gray!20}~DiMP-CA~ 
						& \cellcolor{gray!20}~~26.3~
						& \cellcolor{gray!20}~-~
						& \cellcolor{gray!20}~74.1\color{red}{(2.4$\uparrow$)}~
						& \cellcolor{gray!20}~56.8\color{red}{(7.2$\uparrow$)}~
						& \cellcolor{gray!20}~65.2\color{red}{(4.1$\uparrow$)}~
						\\
						~STARK~ 
						& ~~23.3~
						&~11.5~
						&~76.1~
						&~61.2~
						&~67.2~
						\\
						\cellcolor{gray!20}~STARK-CA~ 
						& ~\cellcolor{gray!20} 23.6~
						& \cellcolor{gray!20} ~8.7\color{blue}{(2.8$\downarrow$)}~
						& \cellcolor{gray!20} ~77.8\color{red}{(1.7$\uparrow$)}~
						&  \cellcolor{gray!20}~62.7\color{red}{(1.5$\uparrow$)}~ 
						&  \cellcolor{gray!20}~68.3\color{red}{(1.1$\uparrow$)}~
						\\
						\midrule[0.05em]
						~\textbf{Pixel-level Tracker}~ &~~~~\textbf{Params(M)}~&~~~\textbf{Flops(G)}~  &~~~$\mathcal{J}$~ &~$\mathcal{F}$~
						& Mean
						\\
						\midrule[0.01em]
						STM~ 
						& ~~24.5~
						&~-~
						&~69.2~
						&~74.0~ 
						&~71.6~

						\\
						
						\cellcolor{gray!20}~STM-CA~
						& \cellcolor{gray!20}~~25.1~
						& \cellcolor{gray!20}~-~
						& \cellcolor{gray!20}~72.8\color{red}{(3.6$\uparrow$)}~
						& \cellcolor{gray!20}~75.6\color{red}{(1.6$\uparrow$)}~ 
						& \cellcolor{gray!20}~74.2\color{red}{(2.6$\uparrow$)}~
						
						\\
						\bottomrule[0.05em]
						
					\end{tabular}
				\end{threeparttable}
			}
		\end{center}
		
		\vspace{-6mm}
		\caption{Improvements of CATs over baselines on GOT-10k~\cite{GOT10K} and DAVIS17~\cite{DAVIS} benchmarks. $\mathcal{J}/\mathcal{F}$ denotes the mean of the region similarity/contour accuracy. }
		\label{tab:improve}
		
		\vspace{-6mm}
	\end{table}

	\vspace{-3mm}
	\paragraph{Testing.}
	For SBT/Siamese, we adopt fixed template as~\cite{dualtfr}. For DCF/STM, the target is cropped as template. The inputs are firstly fused with template by SBT.

	\subsection{Comparison to State-of-the-Art Trackers} \label{sec:Comparison}
	
	\paragraph{GOT-10K.} GOT-10K~\cite{GOT10K} is a large-scale benchmark which has the zero overlap of object classes between training and testing. We follow the official policy without extra training data. As shown in Tab.~\ref{tab:got10k} and Fig.~\ref{fig:otb}, in a fair comparison scenario, our base and large version outperform other top-performing trackers such as STARK-st101, TransT, TrSiam, and DiMP, verifying the strong generalization to unseen objects. Our light and small version also achieve competitive results with much smaller size. 
	\vspace{-1mm}
	\paragraph{OTB100/VOT2020/LaSOT.} We refer the reader to~\cite{OTB2015, vot2020, LaSOT} for detailed descriptions of datasets. 
	In challenging short-term benchmarks~(VOT2020 and OTB100), 
	Tab.~\ref{tab:VOT2020} shows that SBT-small achieves competitive result, which is better than SuperDiMP. After increasing the model variants, SBT-base obtains an EAO of 0.515, being superior to other top-performing trackers. With a much simpler pipeline, our SBT-large is even closed to the winner of VOT2020 challenge RPT~\cite{RPT} (0.530 EAO). Fig.~\ref{fig:otb} shows our base and large version achieves sota results in OTB.
	In long-term benchmark LaSOT, 
	with the comparable model size and no online update, 
	SBT-base outperforms the recent strong Transformer-based methods (STARK-s50 and TransT).

	\subsection{Improvement over Baselines} \label{sec:improvement}
	\vspace{-1mm}
	\paragraph{Box-Level tracking.}
	In Tab.~\ref{tab:improve}, our correlation-aware features improve other tracking pipelines with comparable model size and less computation burden. 
	It validates the generality of our feature network. 
	
	\vspace{-3mm}
	\paragraph{Pixel-Level tracking.}
	In multi-object video object segmentation~(VOS) benchmark DAVIS17~\cite{DAVIS}, STM-CA improves STM by $3.6\%$ in terms of $\mathcal{J}$, proving that VOS methods can benefit from our discriminative embeddings.

	\begin{table}[t]
		\centering
		\begin{center}
			\setlength{\tabcolsep}{0.01mm}
			\fontsize{5.5pt}{2.0mm}\selectfont
			
			\resizebox{1\columnwidth}{!}{
				\begin{threeparttable}
					\begin{tabular}{ @{}c@{} @{}c@{} @{}c@{} @{}c@{} @{}c@{} @{}c@{} @{}c@{} @{}c@{} @{}c@{} @{}c@{} }
						\toprule[0.03em]
						\textbf{Setting}~ &\textbf{Fea.Ext}~&~\textbf{Corr.Emd}~~ &~\textbf{Fea.Corr}~ &~\textbf{Low}& \textbf{Mid}&\textbf{High}~ & ~{AO}~~
						\\
						\midrule[0.01em]
						
						\ding{172}~
						& \cellcolor{gray!5}~ResNet-50~ 
						&~\ding{55}~
						&~DW-Cor~
						&~S2~ &~S3~ &~S4~~
						&\cellcolor{gray!5}~56.2~
						\\
						\ding{173}~ 
						& \cellcolor{gray!10}~ResNet-50~
						&~\ding{55}~
						& ~CA~
						&~S2~ &~S3~ &~S4~~
						&\cellcolor{gray!10}~57.5~
						\\
						\ding{174}~ 
						& \cellcolor{gray!15}~ResNet-50~
						&~\ding{55}~
						&~DCF\tnote{1}~
						&~-~ &~-~ &~S4~~
						&\cellcolor{gray!15}~30.3~
						\\
						\ding{175}~ 
						& \cellcolor{cyan!5}~SBT-base~
						&~\ding{55}~
						&~DW-Corr~
						&~S3B6~ &~S3B8~ &~S3B10~~
						&\cellcolor{cyan!5}~60.1~\color{red}{(3.9$\uparrow$)}
						\\
						\ding{176}~
						&\cellcolor{cyan!10}~SBT-base~
						&~\ding{55}~
						&~CA~
						&~S3B6~ &~S3B8~ &~S3B10~~
						&\cellcolor{cyan!10}~61.5~\color{red}{(4.0$\uparrow$)}
						\\
						\ding{177}~ 
						& \cellcolor{cyan!15}~SBT-base~
						&~\ding{55}~
						&~DCF~
						&~-~ &~-~ &~S3B10~~
						&\cellcolor{cyan!15}~31.5~\color{red}{(1.2$\uparrow$)}
						\\
						\ding{178}~
						& \cellcolor{cyan!5}~SBT-base~
						&\cellcolor{blue!5}~\ding{51}~
						&~-~
						&~S3B6 &~S3B8~ &~S3B10~~
						&\cellcolor{blue!5}~65.0~\color{red}{(3.5$\uparrow$/7.5$\uparrow$)}
						\\
						\ding{179}~ 
						&\cellcolor{cyan!10}~SBT-base~
						&\cellcolor{blue!10}~\ding{51}~
						&~DW-Cor~
						&~S3B6~ &~S3B8~ &~S3B10~~
						&\cellcolor{blue!10}~65.9~\color{red}{(4.8$\uparrow$/9.7$\uparrow$)}
						\\
						\ding{180}~ 
						&\cellcolor{cyan!15}~SBT-base~
						&\cellcolor{blue!15}~\ding{51}~
						&~DCF~
						&~S3B6~ &~S3B8~ &~S3B10~~
						&~35.2\cellcolor{blue!15}~\color{red}{(3.7$\uparrow$/4.9$\uparrow$)}
						\\
						\bottomrule[0.03em]
						
					\end{tabular}
				\end{threeparttable}
			}
		\end{center}
		
		\vspace{-5mm}
		\caption{Ablation studies on GOT-10k~\cite{GOT10K}. S3B6 denotes the third stage $6^{th}$ block. Fea.Ext denotes the feature extraction network. Corr.Emd denotes whether network embeds correlation into extraction layers. Fea.Cor denotes the feature correlation method. For DCF, we integrate SBT-base to ECO~\cite{ECO}. }
		\label{tab:abla}
		
		\vspace{-2mm}
	\end{table}

	\subsection{Exploration Study} \label{sec:Ablation}
	We further explore the characteristics of our SBT feature by training it on the GOT-10k~\cite{GOT10K} training split.

	\begin{figure}[t]
		\centering{\includegraphics[scale = 0.39]{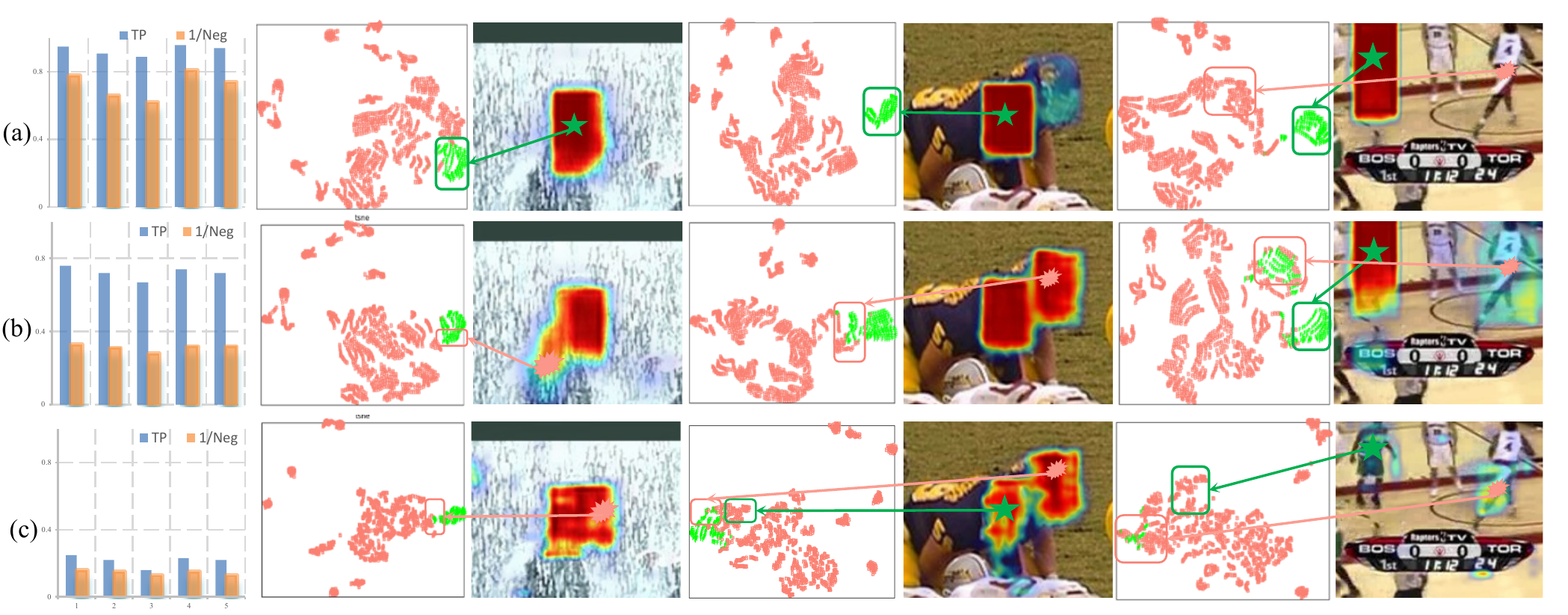}}
		\vspace{-6mm}
		\caption{(a), (b), (c) denote three trackers (refer to Sec.~\ref{sec:discri}). The first sub-figure indicates the average true positive rate and average negative numbers of negative objects. The other sub-figures denote the T-SNE and classification maps.}
		\vspace{-5mm}
		\label{fig:aba_spatial}
	\end{figure}

\begin{figure}[t]
		\centering{\includegraphics[scale = 0.34]{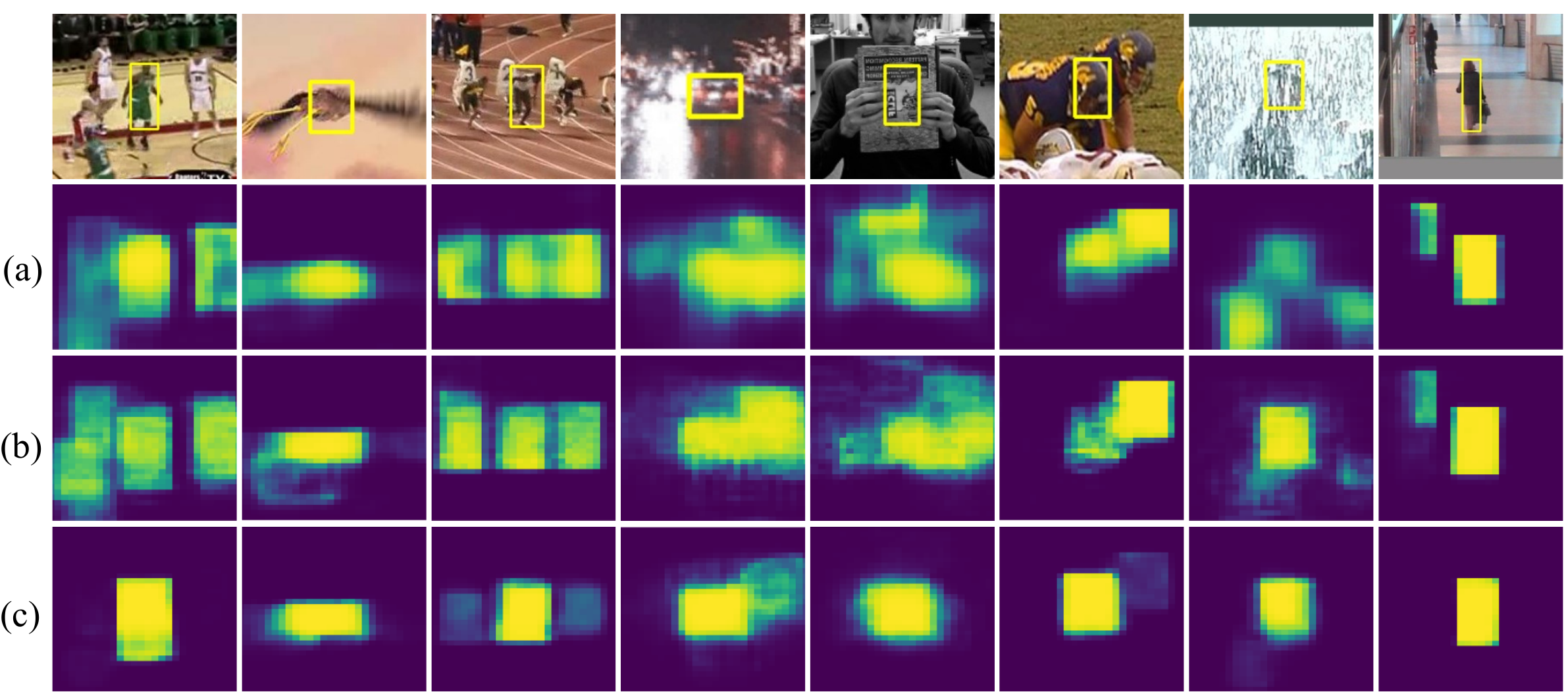}}
		\vspace{-2mm}
		\caption{Visualization of classification (Cls) map ons SBT-base tracker with three different settings. (a): layer-wise aggregation with DW-Corr; (b): layer-wise aggregation with EoC-CA block; (c): correlation-embedded. }
		\vspace{-2mm}
		\label{fig:vis_sea}
	\end{figure}

	\vspace{-1mm}
	\paragraph{Correlation-embedded structure.}
	As shown in Tab.~\ref{tab:abla}, correlation-embedded SBT (\ding{178}, \ding{179}, \ding{180}) significantly improves the tracking performance on all correlation cases (\ding{175}, \ding{176}, \ding{177}).  
	Comparing to the layer-wise aggregation, correlation-embedded trackers outperform the CNN-based trackers or attention-based trackers ($65.9\%$ of \ding{179} Vs. $60.1\%$ of \ding{175}, $65.0\%$ of \ding{178} Vs. $61.5\%$ of \ding{176}, $39.2\%$ of \ding{180} Vs. $30.3\%$ of \ding{174}). 
	It clearly verifies that SBT structure is more effective on multi-level feature utilization.
	We also prove that CA works better than DW-Corr in feature correlation ($60.1\%$ of \ding{175} Vs. $61.5\%$ of \ding{176} ). 
	Fig.~\ref{fig:vis_sea} also shows the superiority of correlation-embedded structure.

	\vspace{-1mm}
	\paragraph{Target-dependent feature embedding.}
	\label{sec:discri}
	We further explore the features of three different settings in two folds: one is to maintain spatial location information while another is to classify the target from distractor objects.
	We begin by training three models with Cls head only to localize the target: 
	(a) Correlation-embedded tracker. 
	(b) Siamese correlation with SBT.
	(c) Siamese correlation with ResNet-50.  
	We select the five hard videos from OTB~\cite{otb} benchmark. 
	The search image randomly jitters around target. We only evaluate the Cls map for localization. 
	In Fig.~\ref{fig:aba_spatial}, the true positive rate of target ground-truth indicates that (a), (b) can preserve more spatial information than CNN (c). The T-SNE/Cls map also show the target-dependent characteristic of (a) features. The average negative objects (largest connected components) of (a) is higher than (b) which indicates that correlation-embedded is critical to filter out distractors.

	\vspace{-1mm}
	\paragraph{Benefits from pre-training.}
	
\begin{figure}[t]
		\centering{\includegraphics[scale = 0.25]{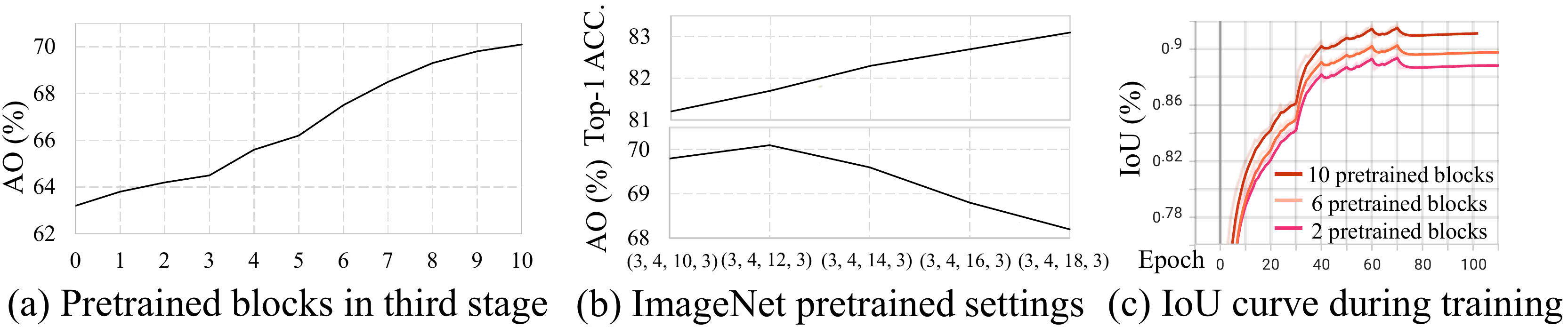}}
		\vspace{-6mm}
		\caption{Tracking performance of SBT with various pretrained settings. (a) denotes the pretrained block number. In (b), different models are to initialize the tracking model. (c) denotes the IoU curves during training epochs. }
		\vspace{-4mm}
		\label{fig:aba_imagenet}
	\end{figure}

	Comparing to the existing trackers~\cite{transt, stark, dualtfr}, our tracking model except prediction heads can be directly benefit from the ImageNet~\cite{ImageNet} pre-trained weights.  
	As shown in Fig.~\ref{fig:aba_imagenet}~(a), there is a significant correlation between the number of pre-trained blocks and tracking performance. 
	We also investigate the impacts of model variants of SBT.  
	In Fig.~\ref{fig:aba_imagenet}~(b), the SBT tracking model prefers consistent block numbers for pre-training. 
	We also observe that SBT converges faster and the stabilized IoU value rises with more pre-trained model weights.

	\section{Conclusion}
	\label{sec:conclusion}
	In this work, we are the first to propose a target-dependent feature network for VOT.
	Our SBT greatly simplifies the tracking pipeline and converges much faster than recent Transformer-based trackers.  
	Then, we conduct a systematic study on SBT tracking both experimentally and theoretically. 
	Moreover, we implement four versions of SBT network which can improve other popular VOT and VOS trackers. 
	Extensive experiments demonstrate that our method achieves sota results and can be applied to other tracking pipelines as dynamic feature network.

	\section{Appendix}
	
	In this supplementary material, we first provide some details mentioned in main text in Sec.A. We then report more exploration studies of our method in Sec.B and provide some visualizations in Sec.C.  
	\section*{A. Experiment Details}
	\label{sec:A}
	
	\subsection*{A.1. Training Details}
	\label{sec:A1}
	
	\noindent \textbf{ImageNet Pre-training.}
	We follow DeiT~\cite{deit, pvt} and apply random cropping, random horizontal flipping, label-smoothing regularization, mixup, CutMix, and random erasing as data augmentations. We initialize the weights with a truncated normal distribution. During training, we employ AdamW with a momentum of $0.9$, a mini-batch size of $128$, and a weight decay of $5$ e-2 to optimize models. The initial learning rate is set to $1$ e-3 and decreases following the cosine schedule. All models are trained for $300$ epochs from scratch on $8$ V100 GPUs. For test in ImageNet~\cite{ImageNet}, we apply a center crop on the validation set, where a $224 \times 224$ patch is cropped to evaluate the classification accuracy.

	\noindent \textbf{Finetune on Tracking.}
	We finetune the whole model on the tracking datasets. In particular, for each pair of search/template images from the training dataset, we compute the losses based on the classification and regression outputs from the prediction head. We use standard cross-entropy loss for the classification loss: all pixels within the ground-truth box are regarded as positive samples and the rest are negative. We use GIoU~\cite{GIoULoss} loss and $L_1$ loss for the regression loss. We load the pre-trained SBT parameters. We use $8$ tesla V100 GPUs and set the batch size to be $20$ for each GPU. Since our model does not have batch normalization, it is not sensitive to the batch size. The batch size can be flexibly adjusted based on the hardware. The search area factor of template and search image is set to 2 and 4, respectively. For GOT-10k training set, the sample pairs of each epoch is $50, 000$. The learning rate is set to be $10^{-4}$ for the feature extraction network, and $10^{-3}$ for the rest. The learning rate decays at the $30_{th}, 50_{th}$ epoch. We finetune the model for $100$ epochs. 
	For full-dataset training, it includes the train subsets of LaSOT~\cite{LaSOT}, GOT-10K~\cite{GOT10K}, COCO2017~\cite{COCO}, and TrackingNet~\cite{trackingnet}. All the forbidden sequences defined by the VOT2019 challenge are abandoned. The pairs of training images in each iteration are sampled from one video sequence or constructed by a static image. On static images, we also construct an image pair by applying data augmentation like flip, brightness jittering and target center jittering.   
	
	\noindent \textbf{Training loss.}
	To validate the generality of our framework, we adopt a vanilla anchor-free prediction head following~\cite{transt} which employs the standard binary cross-entropy loss for classification, which is defined as
	\begin{equation}
	\begin{split}
	\label{equation:BCE_loss}
	\mathcal{L}_{cls} = -\sum_j[y_j{\rm log}(p_j)+(1-y_j){\rm log}(1-p_j)], 
	\end{split}
	\end{equation}
	where $y_j$ denotes the ground-truth label of the $j$-th feature token, $y_j=1$ denotes foreground, and $p_j$ denotes the predicted confidence value belong to  
	the foreground.
	For regression, we apply two kinds of loss: $\ell_1$-norm loss $\mathcal{L}_{1}(.,.)$ 
	and the generalized IoU loss $\mathcal{L}_{GIoU}(.,.)$~\cite{GIoULoss}. 
	The regression loss is as follows:
	\begin{equation}
	\begin{split}
	\label{equation:bbox_loss}
	\mathcal{L}_{reg} = \sum_j\mathbbm{1}_{\{y_j=1\}}[\lambda_{G}\mathcal{L}_{GIoU}(b_j,\hat{b})+\lambda_1\mathcal{L}_1(b_j,\hat{b})]
	\end{split}, 
	\end{equation}
	where $y_j=1$ denotes the positive sample, $b_j$ denotes the $j$-th predicted bounding box, and 
	$\hat{b}$ denotes the normalized ground-truth bounding box.
	We set $\lambda_{G} =5 $ and $\lambda_{1} = 7$ and $12$ for classification loss in our experiments.

	\begin{figure}[t]
		\centering{\includegraphics[scale = 0.27]{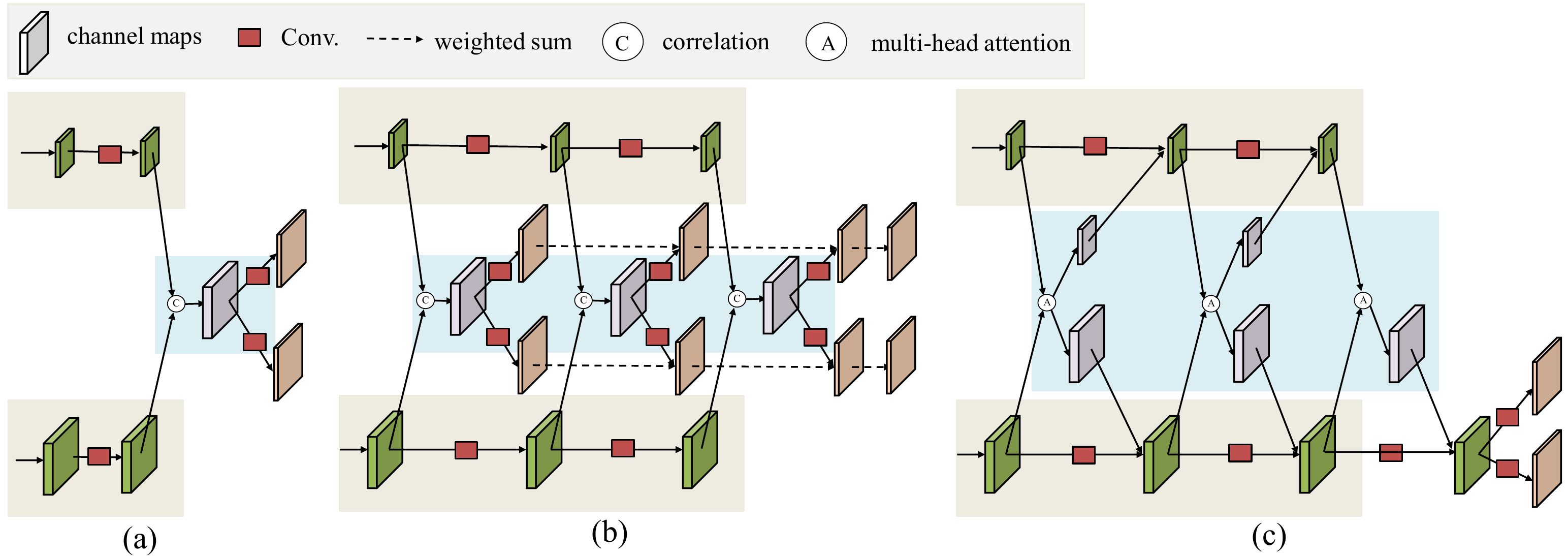}}
		\caption{ (a): a simple Siamese tracking baseline; (b): Siamese tracking baseline with layer-wise aggregation; (c): correlation-embedded structure in SBT. }
		\label{fig:three_pipelines}
	\end{figure}

	\noindent \textbf{Inference Details.}
	For SBT tracker, during inference, the regression head and classification head generate two response maps which embed estimated size shapes and location confidence values, respectively. The maximum confidence value and its bounding box size are chosen to be final predicted target. The template and search image size are chosen to $128 \times 128$ and $256 \times 256$, respectively. To validate the effectiveness of our feature network, no other tricks such as template update and online module are adopted. 

	\noindent \textbf{Ablation Details.}
	In Fig.~\ref{fig:three_pipelines}, 
	It shows the difference among a simple Siamese tracking baseline, Siamese tracking baseline with layer-wise aggregation and correlation-embedded structure in SBT.
	For the Siamese pipeline, the template features are center-cropped with he spatial size  of $7 \times 7$, and then perform correlation with the search features. 
	For cross attention, the template features dose not need to be cropped for the strong global modelling of attention scheme. 
	The channel dimension of the template and search features are adjusted to $256$ by a convolutional layer, which is the same with SiamRPNpp~\cite{siamrpn++}.

	\subsection*{A.2. Correlation-Aware Trackers}
	
	\textbf{SiamFCpp-CA:}
	SiamFCpp~\cite{siamfcpp} is a recent typical Siamese tracker. We replace the GoogLeNet~\cite{googlenet} with the modified SBT-small version. 
	\textbf{DiMP-CA:}
	DiMP~\cite{DiMP} is a modern DCF tracker.. 
	For training and inference, we feed the search and template image pair to modified SBT-base to obtain the correlation-aware search features, then it replaces the original search features extracted by ResNet~\cite{ResNet}.
	\textbf{STARK-CA:}
	STARK~\cite{stark} is a recent strong Transformer-based tracker. 
	We replace the ResNet-50~\cite{ResNet} in STARK with modified SBT-small.
	\textbf{STM-CA:}
	STM~\cite{STM, afb} is a video object segmentation method. We replace the original ResNet-50~\cite{ResNet} to prove that our network can also be used in pixel-wise tracking. 

	\subsection*{A.3. Performance, Model Size and Flops}
	As shown in Fig.~\ref{fig:bab}, we provide a comprehensive comparison in GOT-10k in terms of AO, model size and computation Flops between our methods and other trackers.

	\begin{figure}[t]
		\centering{\includegraphics[scale = 0.30]{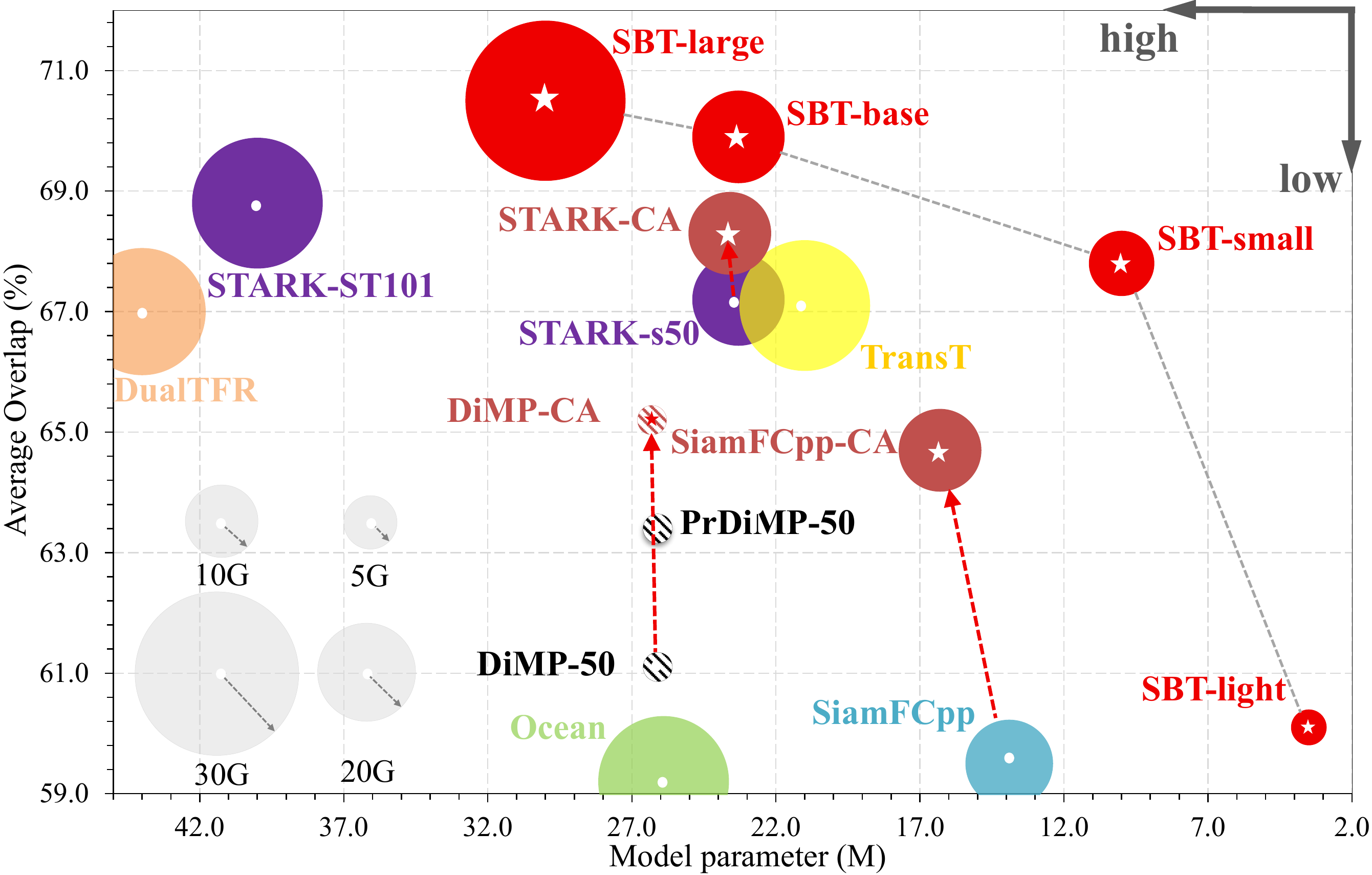}}
		\caption{Comparison to the state-of-the-art methods on the GOT10k dataset. The radius of a circle represents the number of FLOPs of the model. Multiple trackers (with suffix “CA”) can benefit from our correlation-aware features. We do not show the Flops of DCF methods because of the online learning.} 
		\label{fig:bab}
	\end{figure}
	
	\section*{B. More Studies on Model Variants}
	\label{sec:B}
	We further choose the base version of SBT to explore the impacts of model variants on tracking performance.

	\subsection*{B.1. Position Pattern.}
	In Tab.~\ref{tab:position}, we set the total number of EoC-CA block to be $3$ and ablate their position pattern during inference. 
	When the last two correlation operation are set to the $9_{th}$ and $10_{th}$ block, moving the first correlation block to the earlier position achieves better results ($63.2 \%$ of $A_{3}$ vs. $64.6 \%$ of $A_{4}$ vs. $65.7 \%$ of $A_{5}$).  
	It clearly validates that there is a strong correlation between  earliest EoC-CA position and tracking performance. 
	It also points out our key insight that early-cross generates target-dependent features which help tracker to see better.
	From the comparison among $65.0 \%$ of $A_{1}$, $61.0 \%$ of $A_{2}$ and  $63.2 \%$ of $A_{3}$, the interleaved EoC-SA/EoC-CA pattern outperforms the Siamese-style and earlier-cross design. 
	It also enlighten us to choose a interleaved design pattern of SBT tracking. 
	The underlying reason is related to the feature representation: 
	EoC-SA block can refine the template/search feature after the correlation, resulting in a more expressive feature space for matching.
	We also point out that the position pattern can be flexibly designed based on the requirements from future researcher/engineer.

	\begin{table}[t]
		\centering
		\begin{center}
			\setlength{\tabcolsep}{0.01mm}
			\fontsize{5.5pt}{1.0mm}\selectfont
			
			\resizebox{1\columnwidth}{!}{
				\begin{threeparttable}
					\begin{tabular}{ @{}c@{} @{}c@{} @{}c@{} @{}c@{} @{}c@{} @{}c@{} @{}c@{} @{}c@{} @{}c@{} @{}c@{} }
						\toprule[0.03em]
						{Setting}~ &{model}~& ~~~{Pre.}~~ &~{Low}~&~{Mid}~&~~{High}~ & ~~~$\rm SR_{50}$~ & ~$\rm SR_{75}$~ & ~{$\rm  AO$}~
						\\
						\midrule[0.01em]
						~$A_{1}$~ 
						& ~SBT-base~
						&~\ding{51}~
						&~B6~
						&~B8~ &~B10~ 
						&~~~75.3~~
						&~58.6~~
						&\cellcolor{red!30}~~65.0~~
						\\
						~$A_{2}$~
						& ~SBT-base~
						&~\ding{51}~
						&~B8~
						&~B9~ &~B10~ 
						&~~~70.8~~
						&~52.9~~
						&\cellcolor{red!5}~~61.0~~
						\\
						~$A_{3}$~ 
						& ~SBT-base~
						&~\ding{51}~
						&~B6~
						&~B9~ &~B10~ 
						&~~~73.1~~
						&~53.9~~
						&\cellcolor{red!10}~~63.2~~
						\\
						~$A_{4}$~
						& ~SBT-base~
						&~\ding{51}~
						&~B4~
						&~B9~ &~B10~ 
						&~~~75.2~~
						&~58.0~~
						&\cellcolor{red!20}~~64.6~~
						\\
						~$A_{5}$~ 
						& ~SBT-base~
						&~\ding{51}~
						&~B2~
						&~B9~ &~B10~ 
						&~~~75.7~~
						&~59.3~~
						&\cellcolor{red!40}~~65.7~~
						\\
						\bottomrule[0.03em]
						
					\end{tabular}
				\end{threeparttable}
			}
		\end{center}
		
		\vspace{-6mm}
		\caption{Position pattern studies on SBT-base model. Results are from GOT-10k test set. B6 denotes the $6^{th}$ block in third stage. Pre. denotes whether use pre-trained weights or not.}
		\label{tab:position}
		
		\vspace{-4mm}
	\end{table}

	\begin{table}[t]
		\centering
		\begin{center}
			\setlength{\tabcolsep}{0.01mm}
			\fontsize{5.5pt}{1.0mm}\selectfont
			
			\resizebox{1\columnwidth}{!}{
				\begin{threeparttable}
					\begin{tabular}{ @{}c@{} @{}c@{} @{}c@{} @{}c@{} @{}c@{} @{}c@{} @{}c@{} @{}c@{} @{}c@{} @{}c@{} }
						\toprule[0.03em]
						{Setting}~ &{model}~& ~~~{Pre.}~~ &~{N1}~&~~{N2}~&~~{N3}~ & ~~~$\rm SR_{50}$~ & ~$\rm SR_{75}$~ & ~{$\rm  AO$}~
						\\
						\midrule[0.01em]
						~$B_{1}$~ 
						& ~SBT-base~
						&~\ding{55}~
						&~~4~~
						&~~~4~~ &~~12~ 
						&~~~72.3~~
						&~54.9~~
						&\cellcolor{red!20}~~62.3~~
						\\
						~$B_{2}$~
						& ~SBT-base~
						&~\ding{55}~
						&~~4~~
						&~~~4~~ &~~14~ 
						&~~~74.4~~
						&~57.3~~
						&\cellcolor{red!40}~~64.0~~
						\\
						~$B_{3}$~ 
						& ~SBT-base~
						&~\ding{55}~
						&~~2~~
						&~~~2~~ &~~10~ 
						&~~~72.4~~
						&~52.4~~
						&\cellcolor{red!10}~~61.5~~
						\\
						~$B_{4}$~
						& ~SBT-base~
						&~\ding{55}~
						&~2~~
						&~~~2~~ &~~13~ 
						&~~~72.3~~
						&~54.1~~
						&\cellcolor{red!20}~~62.0~~
						\\
						~$B_{5}$~ 
						& ~SBT-base~
						&~\ding{55}~
						&~2~~
						&~~~2~~ &~~15~ 
						&~~~73.4~~
						&~55.7~~
						&\cellcolor{red!30}~~63.1~~
						\\
						\bottomrule[0.03em]
						
					\end{tabular}
				\end{threeparttable}
			}
		\end{center}
		
		\vspace{-6mm}
		\caption{Block pattern studies on GOT-10k test set. N1 denotes the block number in the first model stage. other settings including interleaved CA block and channel dimension are the same. }
		\label{tab:blk}
		
		\vspace{-4mm}
	\end{table}

	\begin{table}[t]
		\centering
		\begin{center}
			\setlength{\tabcolsep}{0.01mm}
			\fontsize{5.5pt}{1.0mm}\selectfont
			
			\resizebox{1\columnwidth}{!}{
				\begin{threeparttable}
					\begin{tabular}{ @{}c@{} @{}c@{} @{}c@{} @{}c@{} @{}c@{} @{}c@{} @{}c@{} @{}c@{} @{}c@{} @{}c@{} }
						\toprule[0.03em]
						{Setting}~ &{model}~& ~~~{Pre.}~~ &~{C1}~&~~{C2}~&~~{C3}~ & ~~~$\rm SR_{50}$~ & ~$\rm SR_{75}$~ & ~{$\rm  AO$}~
						\\
						\midrule[0.01em]
						~$C_{1}$~ 
						& ~SBT-base~
						&~\ding{55}~
						&~64~
						&~~128~ &~~256~ 
						&~~~74.8~~
						&~55.1~~
						&\cellcolor{red!40}~~63.6~~
						\\
						~$C_{2}$~
						& ~SBT-base~
						&~\ding{55}~
						&~32~
						&~~64~ &~~128~ 
						&~~~69.8~~
						&~46.5~~
						&\cellcolor{red!5}~~58.7~~
						\\
						~$C_{3}$~ 
						& ~SBT-base~
						&~\ding{55}~
						&~96~
						&~~192~ &~~384~ 
						&~~~74.3~~
						&~55.3~~
						&\cellcolor{red!40}~~63.7~~
						\\
						~$C_{4}$~
						& ~SBT-base~
						&~\ding{55}~
						&~128~
						&~~256~ &~~512~ 
						&~~~73.0~~
						&~55.2~~
						&\cellcolor{red!30}~~62.8~~
						\\
						~$C_{5}$~ 
						& ~SBT-base~
						&~\ding{55}~
						&~32~
						&~~64~ &~~512~ 
						&~~~72.0~~
						&~51.1~~
						&\cellcolor{red!20}~~61.1~~
						\\
						~$C_{6}$~ 
						& ~SBT-base~
						&~\ding{55}~
						&~64~
						&~~128~ &~~512~ 
						&~~~72.8~~
						&~52.7~~
						&\cellcolor{red!20}~~61.8~~
						\\
						~$C_{7}$~ 
						& ~SBT-base~
						&~\ding{55}~
						&~128~
						&~~256~ &~~512~ 
						&~~~74.1~~
						&~55.8~~
						&\cellcolor{red!40}~~63.6~~
						\\
						\bottomrule[0.03em]
						
					\end{tabular}
				\end{threeparttable}
			}
		\end{center}
		
		\vspace{-6mm}
		\caption{Channel dimension studies on GOT-10k. C1 denotes the channel dimension in the first model stage. other settings including interleaved CA block and block number are the same. }
		\label{tab:channel}
		
		\vspace{-4mm}
	\end{table}

	\subsection*{B.2. Block Number.}
	In Tab.~\ref{tab:blk}, we keep other network factors be the same and ablate their block numbers in different stage. $64.0 \%$ of $B_{2}$ achieves the best tracking performance which has moderate number of blocks ($4/4$) in the shallow model stage and the second large number of blocks ($14$) in the deep stage.  
	It indicates that a moderate size of blocks in shallow stage is necessary. Putting the most of blocks in the third stage is more practical since the spatial size of features is reduced. 
	Obviously, the tracking performance is highly related the size of overall model. 
	Finally, it enlightens us that we can design different versions of SBT trackers to meet the requirements on speed and model size. It is also quite practical and easy to be implemented by stack different number of EoC blocks. 
	
	\subsection*{B.3. Channel Dimension.}
	In Tab.~\ref{tab:blk}, we keep other network factors be the same and ablate their channel dimension in different stage. Since the channel dimension is quite essential to the model size of SBT, it is vital to investigate their impacts on tracking performance. 
	It is natural that larger dimension in each stage can achieve a better performance which also increases the model size ($63.6 \%$ of $C_{1}$, $63.7 \%$ of $C_{3}$ and  $63.6 \%$ of $C_{7}$). 
	It is quite interesting to observe that the gap between shallow stage and deep stage is harmful to the performance. When the shallow stage is set to have $64$ and $128$ channel dimension, then the third stage with $512$ dimension performs worse than that with $256$ dimension($63.6 \%$ of $C_{1}$ vs. $61.8 \%$ of $C_{6}$). 
	It indicates that we should adopt a progressive increment design for the channel dimension in SBT. 
	Moreover, too much channels for the shallow stage encoding seems not helpful but increases the model size a lot ($63.6 \%$ of $C_{1}$ vs.$63.6 \%$ of $C_{7}$).  
	In summary, it is more practical to choose some channel dimension numbers which are widely seen in CNN networks. We can also modify the channel dimensions based on the requirements on different speed and performance.

	\section*{C. Visualization Result}
	
	\begin{figure}[t]
		\centering{\includegraphics[scale = 0.35]{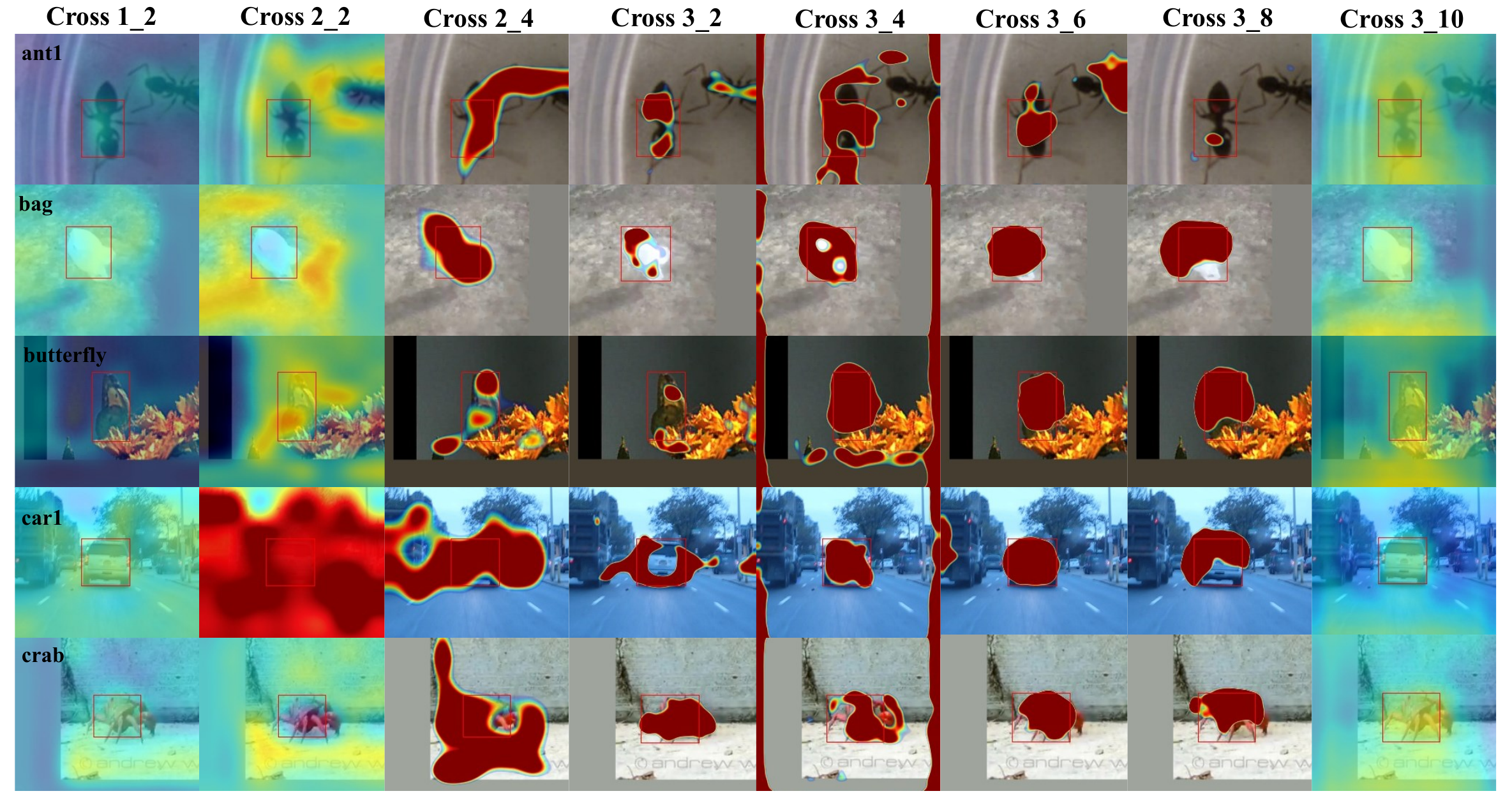}}
		\vspace{-6mm}
		\caption{Cross attention map in each stage of SBT tracker.}
		\vspace{-3mm}
		
		\label{fig:Cross}
	\end{figure}
	
	\begin{figure}[t]
		\centering{\includegraphics[scale = 0.48]{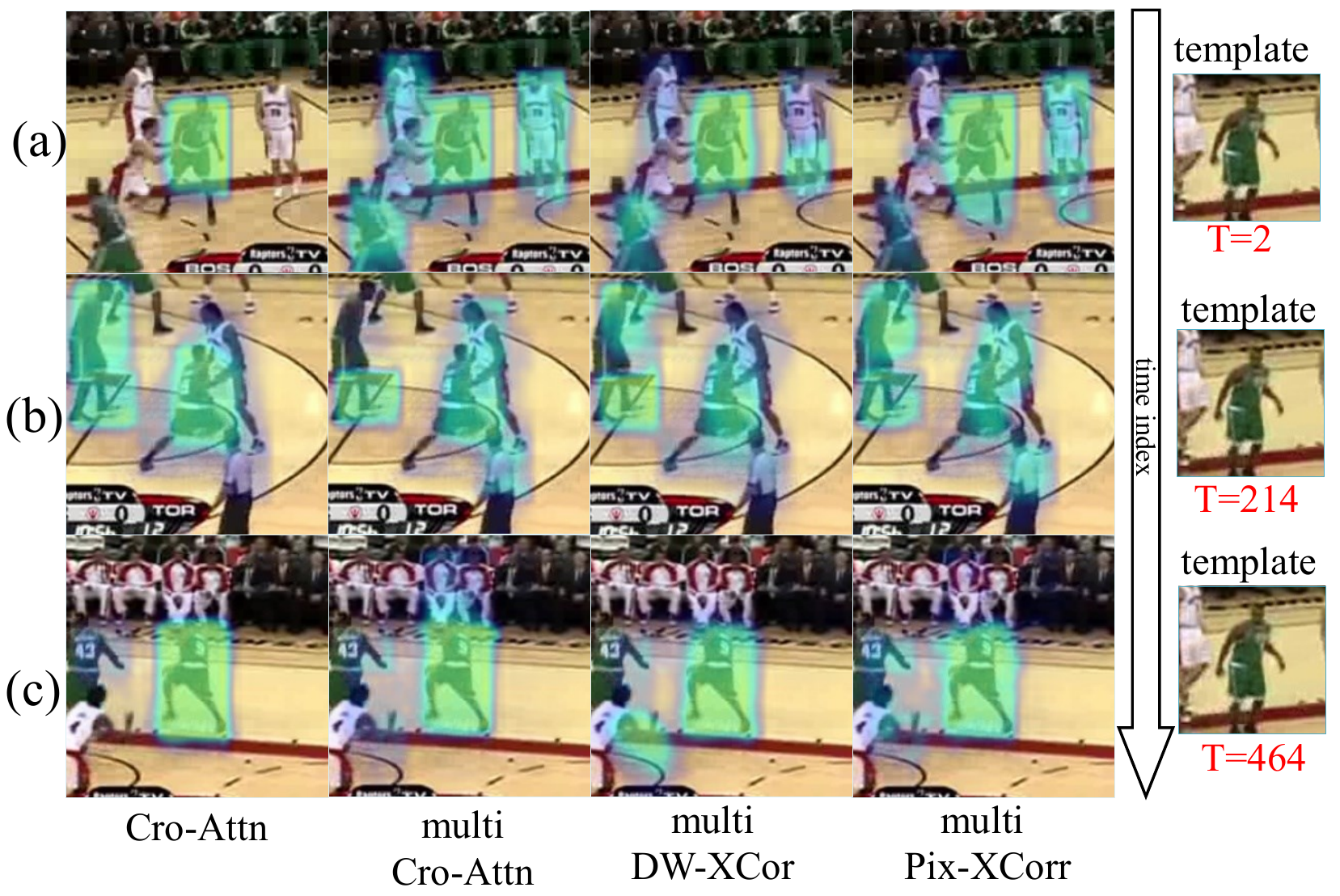}}
		\vspace{-3mm}
		\caption{Visualization of classification map on three cases. (a) denotes suitable template with distractors. (b) denotes template drift with closely attached distractors. (c) denotes template drift with distractors on the edge.}
		\vspace{-5mm}
		\label{fig:vis_four}
	\end{figure}
	
	\begin{figure}[t]
		\centering{\includegraphics[scale = 0.38]{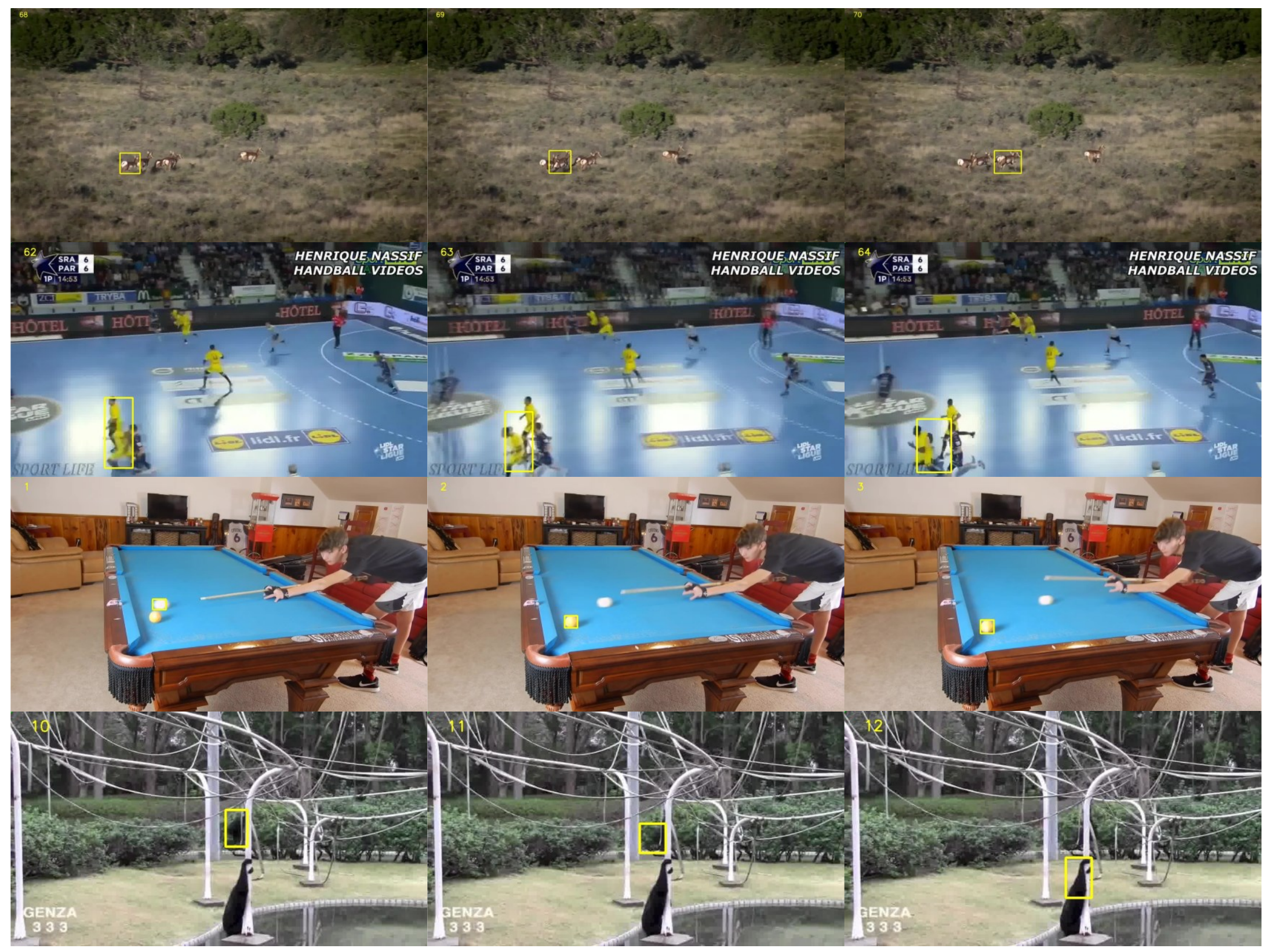}}
		\caption{Failure case. SBT tracker (base) fails on the case of occluded target/out of search region.}
		\label{fig:fail}
	\end{figure}

	\begin{figure*}[t]
		\centering{\includegraphics[scale = 0.55]{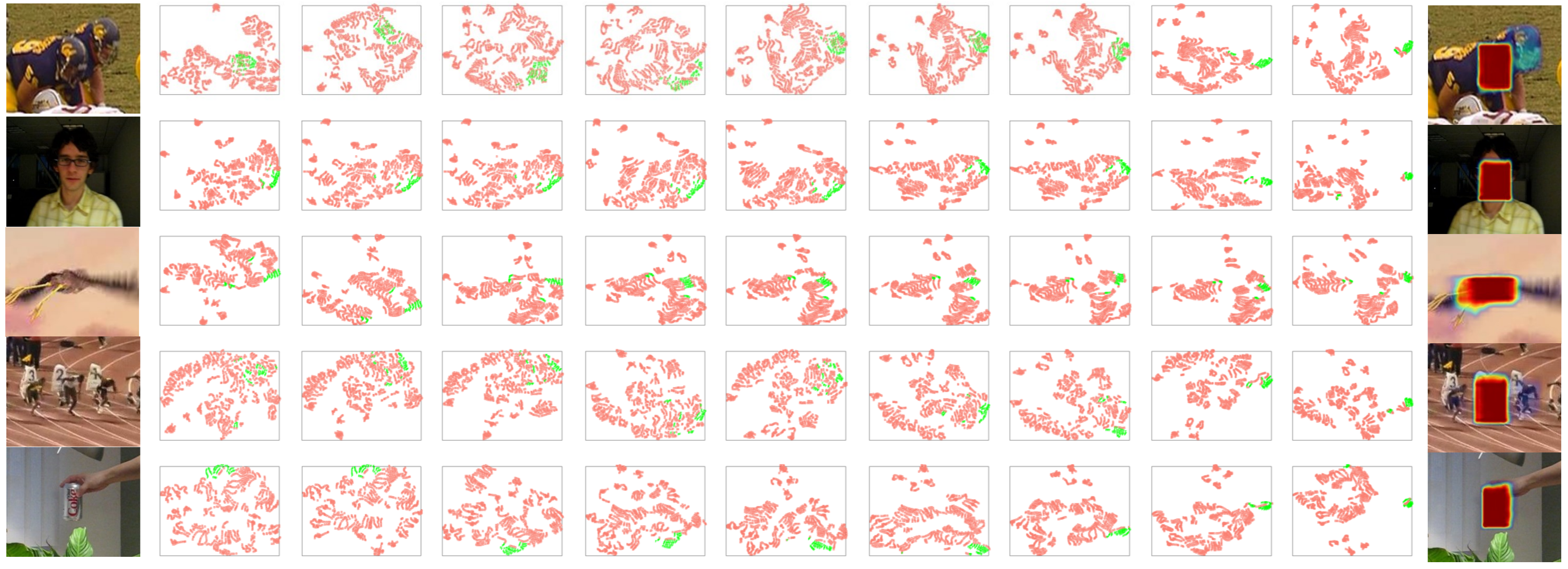}}
		\caption{ TSNE~\cite{tsne} visualizations of search features in correlation-embedded SBT tracker when feature networks go deeper. }
		\label{fig:tsne1}
	\end{figure*}
	
	\begin{figure*}[t]
		\centering{\includegraphics[scale = 0.55]{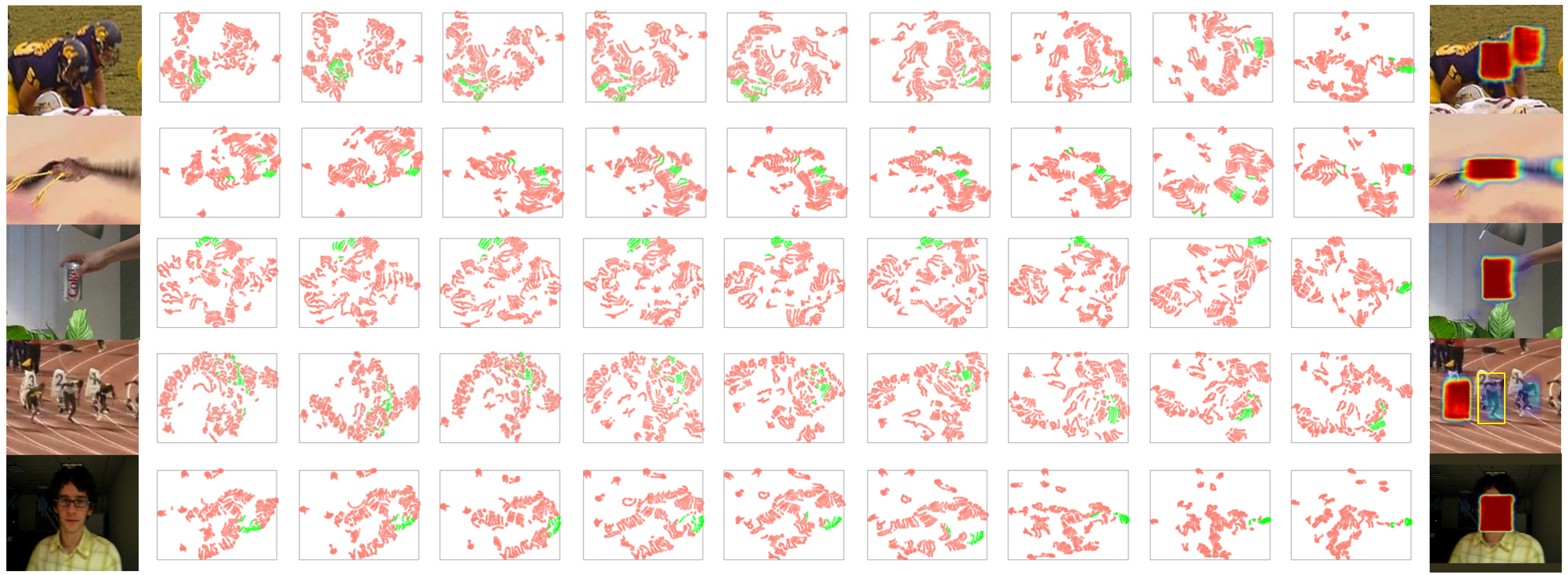}}
		\caption{TSNE~\cite{tsne} visualizations of search features in SBT tracker with Siamese-like extraction when feature networks go deeper.}
		\label{fig:tsne2}
	\end{figure*}
	
	\begin{figure*}[t]
		\centering{\includegraphics[scale = 0.81]{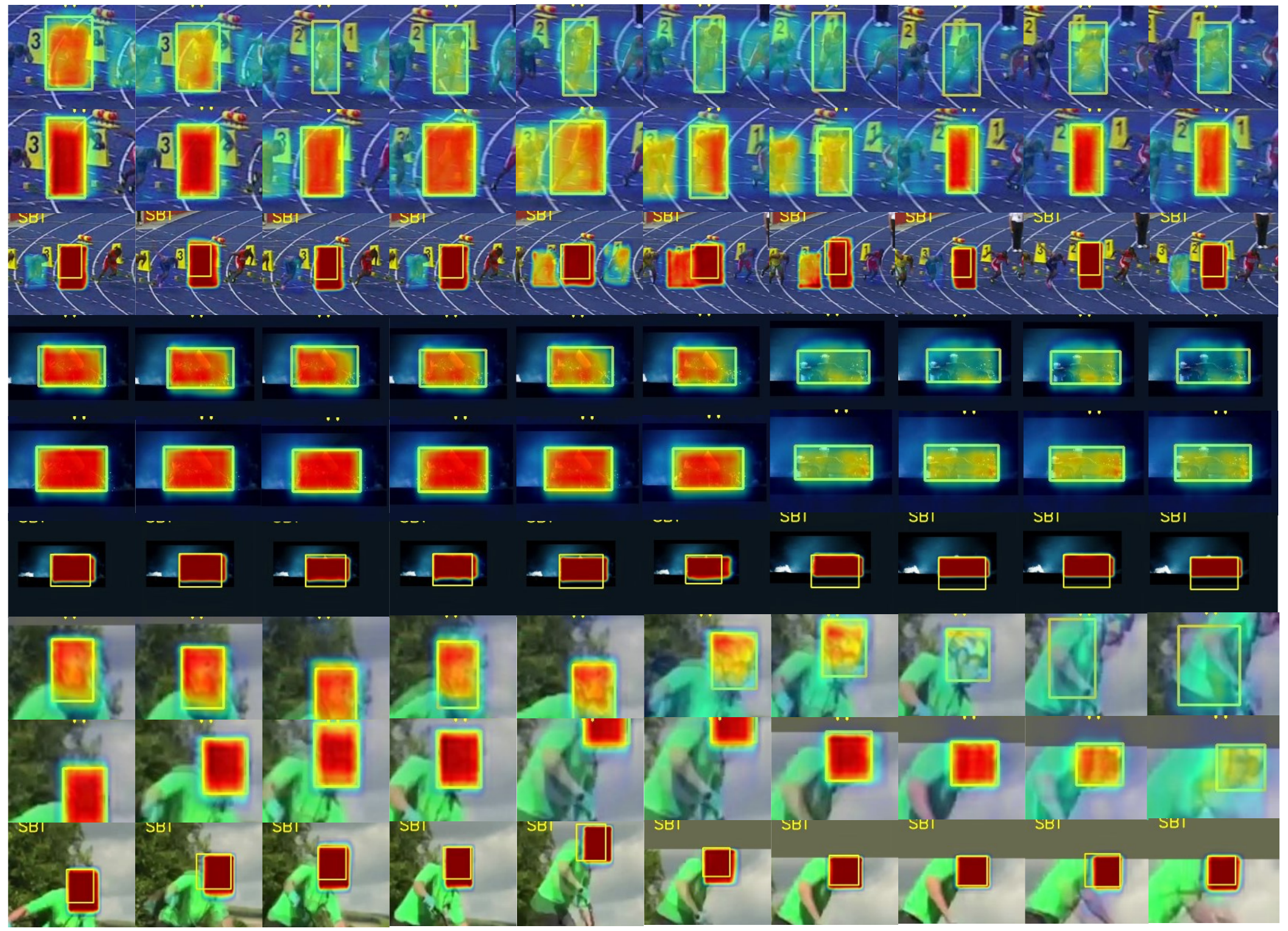}}
		\caption{Visualization tracking results of SiamFCpp (first row), SiamFCpp-CA (second row), our SBT tracker (third row) on the challenging sequences from OTB100. We can see that SBT shows stronger generalization ability and better accuracy throughout tracking. SiamFCpp-CA has more stronger discriminative ability than original SiamFCpp with standard Siamese extraction network towards distractor objects and background clutters. Best viewed with zooming
			in.}
		\label{fig:vis_compare}
	\end{figure*}
	
	\subsection*{C.1. Attention Visualization}
	As shown in Fig.~\ref{fig:Cross}, the attention weights focus on the background context of the search area in the shallow stage. 
	We also vividly observe that it effectively suppresses non-target features in the search image layer by layer. 
	It clearly illustrates that our attention block can discriminate the distractors to some extent.
	In the last block, the attention weight is changed to an uniform distribution which indicates that the search features are ready to the prediction networks. 
	Our Single Branch Transformer (SBT) network allows the features of the two images to deeply interact with each other at the stage of feature extraction which can have dynamic instance-varying behaviors.

	\subsection*{C.2. Response Visualization}
	As shown in Fig.~\ref{fig:vis_four}, we visualize the hard case (basketball video) with numerous distractor objects. 
	Our correlation-aware features can discriminate the distractors in a fine-grained level. 
	When the template drifts, our SBT tracker can also tend to make a more reasonable choices. 
	When time index is 214, the man with white clothes are suppressed in SBT while has higher reponse value in other three models. 
	The distractor object with green clothes is almost the same the target which cannot be identified by human. 
	Thus, it is reasonable to have high response values for the tracker. The other three models perform worse than SBT. 
	When the time index is 464, though the distractor object with green clothes is similar to the target, but can be identified by the white number on the clothes.
	SBT identifys this case successfully while other three fails. It clearly indicates that SBT can have more fine-grained discriminative ability among those appearance-based methods.

	\subsection*{C.3. Failure Case}
	
	When the target object is occluded with distractor objects, together with appearance changes, the pairwise tracking pipeline is hard to figure out the target. 
	It is also commonly seen in many Siamese trackers. 
	Therefore, our framework struggles to
	handle the heavy occlusion (e.g., Fig~\ref{fig:fail}) or out-of-view.
	Another potential limitation of our work is that modern scientific computation packages are not friendly to fast attention computation.

	\subsection*{C.4. TSNE Visualization of Features}
	In Fig.~\ref{fig:tsne1} and Fig.~\ref{fig:tsne2}, we visualize the TSNE of features from our target-dependent network and standard target-unaware Siamese extraction network. When the our SBT network goes deeper, the features belonging to the target (green) become more and more separated from the background and distractors (pink). 
	In the meantime, the search features from the Siamese extraction are totally target-unaware which heavily rely on the separated correlation step to discriminate the targets from background.

	\subsection*{C.5. Visualization of Correlation-Aware Trackers}
	In Fig.~\ref{fig:vis_compare}, we visualize the tracking results of SiamFCpp (first row), SiamFCpp-CA (second row), our SBT tracker (third row) on the challenging sequences from OTB100. 
	We can see that SBT shows stronger discriminative ability and better accuracy throughout tracking. 
	The reponse map of SBT is more centralized and much higher comparing to the background pixels which shows the tracker preserves more spatial information and more discriminative towards the disctractor objects. 
	We can also observe that
	SiamFCpp-CA has more stronger discriminative ability than original SiamFCpp with standard Siamese extraction network towards distractor objects and background clutters. 
	The response map from our correlation-aware features are more discriminative towards background clutters and more suitable for a instance-level task.

	{\small
		\bibliographystyle{ieee_fullname}
		\bibliography{ref}

\begin{thebibliography}{10}\itemsep=-1pt

\bibitem{SuperDiMP}
\href{https://github.com/visionml/pytracking/blob/master/MODEL_ZOO.md}{https://github.com/visionml/pytracking}.

\bibitem{SiameseFC}
Luca Bertinetto, Jack Valmadre, Jo{\~a}o~F Henriques, Andrea Vedaldi, and
  Philip H~S Torr.
\newblock Fully-convolutional siamese networks for object tracking.
\newblock In {\em ECCVW}, 2016.

\bibitem{DiMP}
Goutam Bhat, Martin Danelljan, Luc~Van Gool, and Radu Timofte.
\newblock Learning discriminative model prediction for tracking.
\newblock In {\em ICCV}, 2019.

\bibitem{detr}
Nicolas Carion, Francisco Massa, Gabriel Synnaeve, Nicolas Usunier, Alexander
  Kirillov, and Sergey Zagoruyko.
\newblock End-to-end object detection with transformers.
\newblock In {\em ECCV}, 2020.

\bibitem{transt}
Xin Chen, Bin Yan, Jiawen Zhu, Dong Wang, Xiaoyun Yang, and Huchuan Lu.
\newblock Transformer tracking.
\newblock In {\em CVPR}, pages 8126--8135, 2021.

\bibitem{siamban}
Zedu Chen, Bineng Zhong, Guorong Li, Shengping Zhang, and Rongrong Ji.
\newblock Siamese box adaptive network for visual tracking.
\newblock In {\em CVPR}, 2020.

\bibitem{siamrn}
Siyuan Cheng, Bineng Zhong, Guorong Li, Xin Liu, Zhenjun Tang, Xianxian Li, and
  Jing Wang.
\newblock Learning to filter: Siamese relation network for robust tracking.
\newblock In {\em CVPR}, pages 4421--4431, 2021.

\bibitem{twins}
Xiangxiang Chu, Zhi Tian, Yuqing Wang, Bo Zhang, Haibing Ren, Xiaolin Wei,
  Huaxia Xia, and Chunhua Shen.
\newblock Twins: Revisiting the design of spatial attention in vision
  transformers.
\newblock {\em arXiv preprint arXiv:2104.13840}, 2021.

\bibitem{condPE}
Xiangxiang Chu, Zhi Tian, Bo Zhang, Xinlong Wang, Xiaolin Wei, Huaxia Xia, and
  Chunhua Shen.
\newblock Conditional positional encodings for vision transformers.
\newblock {\em arXiv preprint arXiv:2102.10882}, 2021.

\bibitem{ECO}
Martin Danelljan, Goutam Bhat, Fahad~Shahbaz Khan, and Michael Felsberg.
\newblock {ECO}: Efficient convolution operators for tracking.
\newblock In {\em CVPR}, 2017.

\bibitem{atom}
Martin Danelljan, Goutam Bhat, Fahad~Shahbaz Khan, and Michael Felsberg.
\newblock Atom: Accurate tracking by overlap maximization.
\newblock In {\em CVPR}, 2019.

\bibitem{vit}
Alexey Dosovitskiy, Lucas Beyer, Alexander Kolesnikov, Dirk Weissenborn,
  Xiaohua Zhai, Thomas Unterthiner, Mostafa Dehghani, Matthias Minderer, Georg
  Heigold, Sylvain Gelly, Jakob Uszkoreit, and Neil Houlsby.
\newblock An image is worth 16x16 words: Transformers for image recognition at
  scale, 2020.

\bibitem{LaSOT}
Heng Fan, Liting Lin, Fan Yang, Peng Chu, Ge Deng, Sijia Yu, Hexin Bai, Yong
  Xu, Chunyuan Liao, and Haibin Ling.
\newblock {LaSOT}: A high-quality benchmark for large-scale single object
  tracking.
\newblock In {\em CVPR}, 2019.

\bibitem{CascadedSiameseRPN}
Heng Fan and Haibin Ling.
\newblock Siamese cascaded region proposal networks for real-time visual
  tracking.
\newblock In {\em CVPR}, 2019.

\bibitem{gat}
Dongyan Guo, Yanyan Shao, Ying Cui, Zhenhua Wang, Liyan Zhang, and Chunhua
  Shen.
\newblock Graph attention tracking.
\newblock In {\em CVPR}, pages 9543--9552, 2021.

\bibitem{SiamCAR}
Dongyan Guo, Jun Wang, Ying Cui, Zhenhua Wang, and Shengyong Chen.
\newblock {SiamCAR}: Siamese fully convolutional classification and regression
  for visual tracking.
\newblock In {\em CVPR}, 2020.

\bibitem{ResNet}
Kaiming He, Xiangyu Zhang, Shaoqing Ren, and Jian Sun.
\newblock Deep residual learning for image recognition.
\newblock In {\em CVPR}, 2016.

\bibitem{KCF}
João~F Henriques, Rui Caseiro, Pedro Martins, and Jorge Batista.
\newblock High-speed tracking with kernelized correlation filters.
\newblock In {\em ICVS}, 2008.

\bibitem{GOT10K}
Lianghua Huang, Xin Zhao, and Kaiqi Huang.
\newblock {GOT-10k}: A large high-diversity benchmark for generic object
  tracking in the wild.
\newblock {\em TPAMI}, 2019.

\bibitem{got}
Lianghua Huang, Xin Zhao, and Kaiqi Huang.
\newblock Got-10k: A large high-diversity benchmark for generic object tracking
  in the wild.
\newblock {\em IEEE Transactions on Pattern Analysis and Machine Intelligence},
  2019.

\bibitem{vot2020}
Matej Kristan, Ale{\v{s}} Leonardis, Ji{\v{r}}{\'\i} Matas, Michael Felsberg,
  Roman Pflugfelder, Joni-Kristian K{\"a}m{\"a}r{\"a}inen, Martin Danelljan,
  Luka~{\v{C}}ehovin Zajc, Alan Luke{\v{z}}i{\v{c}}, Ondrej Drbohlav, et~al.
\newblock The eighth visual object tracking vot2020 challenge results.
\newblock In {\em ECCVW}, 2020.

\bibitem{siamrpn++}
Bo Li, Wei Wu, Qiang Wang, Fangyi Zhang, Junliang Xing, and Junjie Yan.
\newblock Siamrpn++: Evolution of siamese visual tracking with very deep
  networks.
\newblock In {\em CVPR}, 2019.

\bibitem{siamrpn}
Bo Li, Junjie Yan, Wei Wu, Zheng Zhu, and Xiaolin Hu.
\newblock High performance visual tracking with siamese region proposal
  network.
\newblock In {\em CVPR}, pages 8971--8980, 2018.

\bibitem{targetaware}
Xin Li, Chao Ma, Baoyuan Wu, Zhenyu He, and Ming-Hsuan Yang.
\newblock Target-aware deep tracking.
\newblock In {\em CVPR}, pages 1369--1378, 2019.

\bibitem{afb}
Yongqing Liang, Xin Li, Navid Jafari, and Qin Chen.
\newblock Video object segmentation with adaptive feature bank and
  uncertain-region refinement.
\newblock {\em NIPS}, 2020.

\bibitem{COCO}
Tsung-Yi Lin, Michael Maire, Serge~J. Belongie, Lubomir~D. Bourdev, Ross~B.
  Girshick, James Hays, Pietro Perona, Deva Ramanan, Piotr Doll{\'a}r, and
  C.~Lawrence Zitnick.
\newblock {Microsoft COCO}: Common objects in context.
\newblock In {\em ECCV}, 2014.

\bibitem{swin}
Ze Liu, Yutong Lin, Yue Cao, Han Hu, Yixuan Wei, Zheng Zhang, Stephen Lin, and
  Baining Guo.
\newblock Swin transformer: Hierarchical vision transformer using shifted
  windows.
\newblock {\em arXiv preprint arXiv:2103.14030}, 2021.

\bibitem{AdamW}
Ilya Loshchilov and Frank Hutter.
\newblock Decoupled weight decay regularization.
\newblock {\em arXiv preprint arXiv:1711.05101}, 2017.

\bibitem{RPT}
Ziang Ma, Linyuan Wang, Haitao Zhang, Wei Lu, and Jun Yin.
\newblock Rpt: Learning point set representation for siamese visual tracking.
\newblock {\em arXiv preprint arXiv:2008.03467}, 2020.

\bibitem{trackingnet}
Matthias Muller, Adel Bibi, Silvio Giancola, Salman Alsubaihi, and Bernard
  Ghanem.
\newblock Trackingnet: A large-scale dataset and benchmark for object tracking
  in the wild.
\newblock In {\em ECCV}, 2018.

\bibitem{STM}
Seoung~Wug Oh, Joon-Young Lee, Ning Xu, and Seon~Joo Kim.
\newblock Video object segmentation using space-time memory networks.
\newblock In {\em ICCV}, 2019.

\bibitem{DAVIS}
Federico Perazzi, Jordi Pont-Tuset, Brian McWilliams, Luc Van~Gool, Markus
  Gross, and Alexander Sorkine-Hornung.
\newblock A benchmark dataset and evaluation methodology for video object
  segmentation.
\newblock In {\em CVPR}, pages 724--732, 2016.

\bibitem{GIoULoss}
Hamid Rezatofighi, Nathan Tsoi, JunYoung Gwak, Amir Sadeghian, Ian Reid, and
  Silvio Savarese.
\newblock Generalized intersection over union: A metric and a loss for bounding
  box regression.
\newblock In {\em CVPR}, 2019.

\bibitem{ImageNet}
Olga Russakovsky, Jia Deng, Hao Su, Jonathan Krause, Sanjeev Satheesh, Sean Ma,
  Zhiheng Huang, Andrej Karpathy, Aditya Khosla, and Michael Bernstein.
\newblock {ImageNet} {Large} scale visual recognition challenge.
\newblock {\em IJCV}, 2015.

\bibitem{googlenet}
Christian Szegedy, Wei Liu, Yangqing Jia, Pierre Sermanet, Scott Reed, Dragomir
  Anguelov, Dumitru Erhan, Vincent Vanhoucke, and Andrew Rabinovich.
\newblock Going deeper with convolutions.
\newblock In {\em CVPR}, 2015.

\bibitem{imagetrain}
Hugo Touvron, Matthieu Cord, Matthijs Douze, Francisco Massa, Alexandre
  Sablayrolles, and Herv{\'e} J{\'e}gou.
\newblock Training data-efficient image transformers \& distillation through
  attention.
\newblock In {\em International Conference on Machine Learning}, pages
  10347--10357. PMLR, 2021.

\bibitem{deit}
Hugo Touvron, Matthieu Cord, Matthijs Douze, Francisco Massa, Alexandre
  Sablayrolles, and Herv{\'e} J{\'e}gou.
\newblock Training data-efficient image transformers \& distillation through
  attention.
\newblock In {\em International Conference on Machine Learning}, pages
  10347--10357. PMLR, 2021.

\bibitem{tsne}
Laurens Van~der Maaten and Geoffrey Hinton.
\newblock Visualizing data using t-sne.
\newblock {\em Journal of machine learning research}, 9(11), 2008.

\bibitem{vaswani2017attention}
Ashish Vaswani, Noam Shazeer, Niki Parmar, Jakob Uszkoreit, Llion Jones,
  Aidan~N Gomez, {\L}ukasz Kaiser, and Illia Polosukhin.
\newblock Attention is all you need.
\newblock In {\em NIPS}, 2017.

\bibitem{MAML-track}
Guangting Wang, Chong Luo, Xiaoyan Sun, Zhiwei Xiong, and Wenjun Zeng.
\newblock Tracking by instance detection: A meta-learning approach.
\newblock In {\em CVPR}, 2020.

\bibitem{spm}
Guangting Wang, Chong Luo, Zhiwei Xiong, and Wenjun Zeng.
\newblock Spm-tracker: Series-parallel matching for real-time visual object
  tracking.
\newblock In {\em CVPR}, 2019.

\bibitem{tmt}
Ning Wang, Wengang Zhou, Jie Wang, and Houqiang Li.
\newblock Transformer meets tracker: Exploiting temporal context for robust
  visual tracking.
\newblock In {\em CVPR}, pages 1571--1580, 2021.

\bibitem{attnsiam}
Qiang Wang, Zhu Teng, Junliang Xing, Jin Gao, Weiming Hu, and Stephen Maybank.
\newblock Learning attentions: residual attentional siamese network for high
  performance online visual tracking.
\newblock In {\em CVPR}, pages 4854--4863, 2018.

\bibitem{SiamMask}
Qiang Wang, Li Zhang, Luca Bertinetto, Weiming Hu, and Philip H.~S. Torr.
\newblock Fast online object tracking and segmentation: {A} unifying approach.
\newblock In {\em CVPR}, 2019.

\bibitem{pvt}
Wenhai Wang, Enze Xie, Xiang Li, Deng-Ping Fan, Kaitao Song, Ding Liang, Tong
  Lu, Ping Luo, and Ling Shao.
\newblock Pyramid vision transformer: A versatile backbone for dense prediction
  without convolutions, 2021.

\bibitem{otb}
Yi Wu, Jongwoo Lim, and Ming-Hsuan Yang.
\newblock Online object tracking: A benchmark.
\newblock In {\em CVPR}, 2013.

\bibitem{OTB2015}
Yi Wu, Jongwoo Lim, and Ming~Hsuan Yang.
\newblock Object tracking benchmark.
\newblock {\em IEEE Transactions on Pattern Analysis and Machine Intelligence},
  37(9):1834--1848, 2015.

\bibitem{dualtfr}
Fei Xie, Chunyu Wang, Guangting Wang, Yang Wankou, and Wenjun Zeng.
\newblock Learning tracking representations via dual-branch fully transformer
  networks.
\newblock In {\em ICCVW}, 2021.

\bibitem{samn}
Fei Xie, Wankou Yang, Bo Liu, Kaihua Zhang, Guangting Wang, and Wangmeng Zuo.
\newblock Learning spatio-appearance memory network for high-performance visual
  tracking.
\newblock {\em ICCVW}, 2021.

\bibitem{resnext}
Saining Xie, Ross Girshick, Piotr Doll{\'a}r, Zhuowen Tu, and Kaiming He.
\newblock Aggregated residual transformations for deep neural networks.
\newblock In {\em CVPR}, 2017.

\bibitem{siamfcpp}
Yinda Xu, Zeyu Wang, Zuoxin Li, Ye Yuan, and Gang Yu.
\newblock {SiamFC++:} towards robust and accurate visual tracking with target
  estimation guidelines.
\newblock In {\em AAAI}, 2020.

\bibitem{stark}
Bin Yan, Houwen Peng, Jianlong Fu, Dong Wang, and Huchuan Lu.
\newblock Learning spatio-temporal transformer for visual tracking.
\newblock {\em ICCV}, 2021.

\bibitem{AlphaRefine}
Bin Yan, Xinyu Zhang, Dong Wang, Huchuan Lu, and Xiaoyun Yang.
\newblock Alpha-refine: Boosting tracking performance by precise bounding box
  estimation.
\newblock {\em CVPR}, 2021.

\bibitem{dtt}
Bin Yu, Ming Tang, Linyu Zheng, Guibo Zhu, Jinqiao Wang, Hao Feng, Xuetao Feng,
  and Hanqing Lu.
\newblock High-performance discriminative tracking with transformers.
\newblock In {\em ICCV}, 2021.

\bibitem{transdcf}
Bin Yu, Ming Tang, Linyu Zheng, Guibo Zhu, Jinqiao Wang, Hao Feng, Xuetao Feng,
  and Hanqing Lu.
\newblock High-performance discriminative tracking with transformers.
\newblock In {\em ICCV}, pages 9856--9865, 2021.

\bibitem{deforsiam}
Yuechen Yu, Yilei Xiong, Weilin Huang, and Matthew~R Scott.
\newblock Deformable siamese attention networks for visual object tracking.
\newblock In {\em CVPR}, pages 6728--6737, 2020.

\bibitem{UpdateNet}
Lichao Zhang, Abel Gonzalez-Garcia, Joost van~de Weijer, Martin Danelljan, and
  Fahad~Shahbaz Khan.
\newblock Learning the model update for siamese trackers.
\newblock In {\em ICCV}, 2019.

\bibitem{rest}
Qinglong Zhang and Yubin Yang.
\newblock Rest: An efficient transformer for visual recognition.
\newblock {\em arXiv preprint arXiv:2105.13677}, 2021.

\bibitem{automatch}
Zhipeng Zhang, Yihao Liu, Xiao Wang, Bing Li, and Weiming Hu.
\newblock Learn to match: Automatic matching network design for visual
  tracking.
\newblock 2021.

\bibitem{siamdw}
Zhipeng Zhang and Houwen Peng.
\newblock Deeper and wider siamese networks for real-time visual tracking.
\newblock In {\em CVPR}, 2019.

\bibitem{Ocean}
Zhipeng Zhang, Houwen Peng, Jianlong Fu, Bing Li, and Weiming Hu.
\newblock Ocean: Object-aware anchor-free tracking.
\newblock In {\em ECCV}, 2020.

\bibitem{DCFST}
Linyu Zheng, Ming Tang, Yingying Chen, Jinqiao Wang, and Hanqing Lu.
\newblock Learning feature embeddings for discriminant model based tracking.
\newblock In {\em ECCV}, 2020.

\bibitem{drol}
Jinghao Zhou, Peng Wang, and Haoyang Sun.
\newblock Discriminative and robust online learning for siamese visual
  tracking.
\newblock In {\em AAAI}, volume~34, pages 13017--13024, 2020.

\end{thebibliography}
	}
	
\end{document}